\pdfoutput=1
\documentclass[letterpaper, 10 pt, conference]{ieeeconf}  

\IEEEoverridecommandlockouts                              

\usepackage{amsmath} 
\usepackage{amssymb}  
\usepackage{fancyhdr,graphicx,amsmath,amssymb, xcolor}
\usepackage{amsmath,amssymb,amsfonts}
\usepackage{svg}
\usepackage[ruled,vlined]{algorithm2e}
\usepackage{courier}
\include{pythonlisting}

\usepackage{float}

\DeclareMathOperator*{\argmax}{argmax}

 \usepackage[font=small,skip=2pt]{caption}

\usepackage{url}
\usepackage{pdfpages}

\title{\LARGE \bf
Non-Parametric Stochastic Policy Gradient with Strategic Retreat for Non-Stationary Environment

}

\author{Apan Dastider and Mingjie Lin
}

\begin{document}
\maketitle
\thispagestyle{empty}
\pagestyle{empty}

\begin{abstract}

%


    In modern robotics,
    effectively computing 
    optimal control policies under dynamically varying environments
    poses substantial challenges to 
    the off-the-shelf parametric policy gradient methods,
    such as the Deep Deterministic Policy Gradient (DDPG) and Twin Delayed Deep Deterministic policy gradient (TD3). 
    In this paper, 
    we propose a systematic methodology to dynamically learn
    a sequence of optimal control policies {\em non-parametrically},
    while autonomously adapting with
    the constantly changing environment dynamics.
    Specifically, 
    our non-parametric kernel-based methodology 
    embeds a policy distribution as the features in a non-decreasing Euclidean
    space, therefore allowing its search space to be defined as a very high (possible
    infinite) dimensional RKHS (Reproducing Kernel Hilbert Space).
    Moreover, by leveraging the similarity metric computed in RKHS, 
    we augmented our non-parametric learning with 
    the technique of \texttt{AdaptiveH}--- adaptively
    selecting a time-frame window of finishing the optimal part of whole
    action-sequence sampled on some preceding observed state.
    To validate our proposed
    approach, we conducted extensive experiments with 
    multiple classic benchmarks and one simulated robotics benchmark equipped with dynamically changing environments.
    Overall, our methodology has outperformed the well-established DDPG and TD3 methodology
    by a sizeable margin in terms of learning performance.

%
%

\end{abstract}

\section{INTRODUCTION}

Reinforcement Learning (RL) has revolutionized the process of learning optimal
action mapping policies in diverse problem domains 
ranging from video games\cite{atari} to robotic navigation\cite{how_to_train}. 
Among many Deep Reinforcement Learning (DRL) algorithms,
DDPG (Deep Deterministic Policy Gradient)~\cite{Lillicrap2016} and its upgraded variant TD3 (Twin Delayed Deep Deterministic Policy Gradient) \cite{td3},
have attained immense successes in robotics
through
strategically combining optimal
deterministic policy search algorithms and value-function learning.
%
%
Despite many successes of DRL in robotic learning, 
the vast majority of these studies, however, restrictively assume 
the model of their operating environments to be static.
In fact, while extensive research 
has been conducted 
on policy gradient methods 
in order to
handle real-world high dimensional continuous space environments, 
much less attention has been dedicated to formulate algorithms to handle non-stationary
systems. 

Unfortunately, real-world environments in robotics are
typically nonstationary and evolving, 
therefore governed by dynamically changing transition dynamics 
due to some totally imperceptible causes. 
As such, 
learning optimal control policy under dynamically varying environments 
with conventional methodologies, such as DDPG or TD3,
becomes both computationally intensive and quite time-consuming. 
%
Specifically,
while interplaying with a dynamically {\em self-altering} environment, actions
suggested by a deterministic policy based on the preceding state observation
become outdated as the state-space is continuously evolving. On the contrary,
by learning a stochastic policy in a continuously evolving environment, more
probable actions can be sampled from the policy. Moreover, in case of
deterministic way of learning, it is assumed that a stationary transition
dynamics and reward distribution exist in the environment \cite{Lillicrap2016,
silver2014deterministic}. As a result, deterministic policy gradient method performs suboptimally, when the
learning algorithm is exposed to problems where transition functions and reward
distributions are changing in a non-deterministic fashion. Comparatively, a
probabilistic approach for action sampling from a policy distribution can
suggest actions compatible to handle the stochasticity of the dynamic
environment. In particular, probabilistic algorithms for handling optimization
problems or perception learning often exhibit robustness to sensory
irregularities, random perturbations and stochastic nature of environments and
scale much better to the complex environments, where the agent need to handle
unknown uncertainty in much robust and adaptive way \cite{Thrun_2000}.
%
%

To overcome all these thorny challenges fundamentally caused by 
the non-stationarity of underlying environment, 
we formulate the scheme of
adaptively learning sequence of action-mapping functionals 
by embedding the functionals into a vector-valued kernel space instead of a fixed parametric
space and representing policy distributions defined over a high (possible
infinite) dimensional RKHS~\cite{pmlr-v38-lever15}. 
In this paper, we develop a new class of non-parametric policy gradient methods~\cite{PGT_NSMDP}
which optimally ensures a gradient ascending direction while optimizing and
adaptively executing such actions compatible with the concurrent
dynamics of the environment. 
We validate both theoretically and experimentally that our proposed approach
can handle non-stationary environments with hidden, but bounded, degree of dynamic evolution. In most cases, such
an upper-bound on the non-stationary transition dynamics is reasonably ensured
by considering that the evolution rate of the non-stationary MDPs are bounded
by Lipschitz Continuity assumption\cite{PGT_NSMDP},\cite{worstcase_nsmdp}.
%

{\em Why Non-Parametric Reinforcement Learning?} ---
In a real-world robotic setting, policy computation and policy inference
possess certain latency and it is impossible to pause the environment in the
continuous stream of operating phase. 
So, an adaptive learning framework is highly required to handle
dynamically varying transition distribution and reward function through
updating policy distribution in a more frequent policy update fashion. 
But unfortunately, in parametric setting like DDPG, finding a gradient similarity
metric to find an adaptive termination point for frequent policy updates is
mathematically very difficult and often gradients can not be easily estimated,
which can be easily done in an RKHS based non-parametric learning
framework\cite{pmlr-v38-lever15}, \cite{PGT_NSMDP}. Parametric methods also
impose two major disadvantages. Firstly, for a parametric model like DDPG, it
is very difficult 
to initialize a suitable prior parameter
matrix as picking too large matrix can be computationally expensive or picking
too small will be unable to fit the complex policies need to learn for action
mapping \cite{pmlr-v38-lever15}. Secondly, when the dimension of problem goes
higher, the policy update in algorithm like DDPG grows exponentially. Such
update can cause exploding gradients while dealing with unbounded system
variances. To address theses issues, non-parametric policy learning leverages
transition distributions as kernel embeddings in an RKHS whose complexity
remains linear with dimensionality of problem space.  Non-parametric
representation of control policies can be adaptive as RKHS based kernel trick
enables the agent to learn complex, but sufficient, representation of stochastic control policies when required.

{\em Why Strategic Retreats or Adaptive-H?} ---
Unfortunately, kernel based non-parametric learning possesses one major
disadvantage. Non-parametric method requires a lot of training data to
accurately model the action-mapping functions. RL itself is a data-hungry
method. As a result, the kernel matrix, often stated as Gram Matrix\cite{gram}
becomes huge in dimension and, in turn, brings the problem of {\em
memory-explosion}. Thus, the learning rate for optimization becomes very slow
and kernel centers in the kernel dictionary increase exponentially with more
experiences from the environment. Since, a part of sampled
action-trajectory becomes obsolete as the state-space evolves, the problem of memory-explosion can be prevented by
adaptively presetting a termination trigger to stop {\em non-optimal} action
execution in an episodic trajectory and perform policy update. Thus, the algorithm can be made truly
scalable for very large dimensional problems and can be sample-efficient by truncating the full trajectories at optimal timing.
We term this new formulation in accordance with a non-parametric policy
gradient as \texttt{AdaptiveH} 

{\em Why Combining RKHS Method and Strategic Retreats?} ---
The policy gradients with respect to action-mapping mean functional can be
easily and efficiently estimated in an RKHS following Fr\'echet
Derivative\cite{pmlr-v38-lever15}. Such gradient estimation is not
mathematically sound in parametric TD3 and  DDPG method. Respectively, since policy
distributions are considered as elements of an RKHS and RKHS itself inherits all properties of a vector
space, a dot product based similarity metric \cite{PGT_NSMDP} can be
efficiently utilized to guide the policy search method. By adopting this
similarity metric between consecutive gradient updates, our agent adaptively
develop sequence of control polices in a completely non-stationary setting with
potentially bounded variance in system dynamics.
Moreover, we mention the theoretical convergence guarantee to the fixed point
solution by proving that dynamic regret term becomes a sub-martingale through
continual policy learning. For reward maximization, policy
gradients should move into an ascent
direction during the optimization process to reach global \textit{maxima}
point. Formulated on this concept, we consecutively measure the dot product
between two gradient updates. If the result of the dot product diverges from
being a positive number, the following gradients are deviating from an ascent
direction. At that instance, our agent pre-maturely aborts action-execution and
upgrades itself to a new policy. Thus, we adaptively tune a action-execution
window $H$ after which less important actions are not fully finished. Such
adjustable and frequent policy update brings about sequence of optimal control
policies and this scheme can not be done easily in deterministic policy search
method. 

{\bf Contributions :}
\begin{itemize}

    \item We proposed a non-parametric policy gradient framework which can
        adaptively generate sequence of control policies to handle dynamically
        varying real-world scenarios.

    \item We introduced both theoretical guarantees and experimental proofs
        that by premature action stopping and frequent policy update through a
        dot-product based similarity metric in RKHS, our proposed approach can
        speed up the learning process and enhance sample-efficiency.

    \item We also validate that our proposed approach can exhibit better
        performance than state-of-the-art parametric DRL algorithms like TD3 and DDPG
        in non-stationary environments. 

\end{itemize}

\begin{figure*}[htbp]
    \includegraphics[width=1\textwidth]{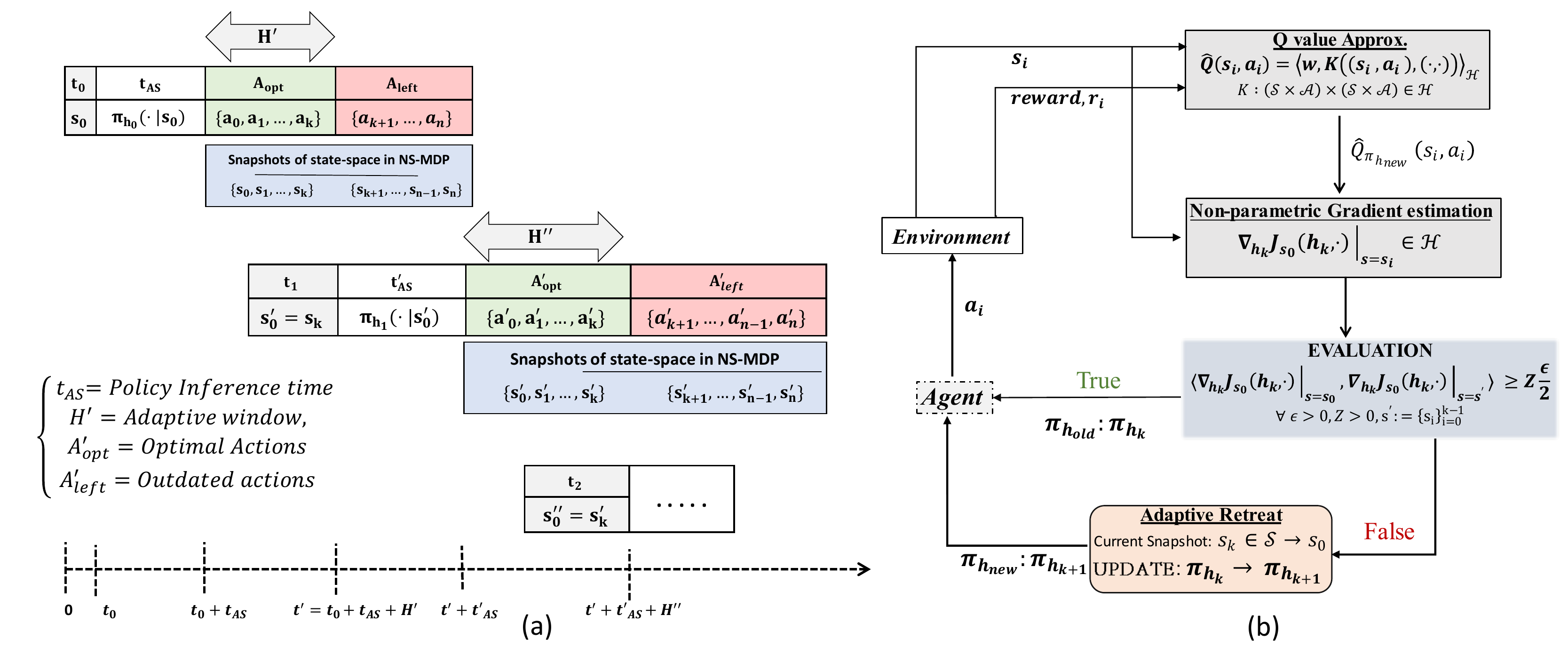}
    \caption{(a) Sketch of optimal Action Selection by estimating Adaptive-H period window and sequence of policy updates. (b) Overview of Algorithmic steps for inner-product based gradient ascent direction assurance for maximizing expected rewards.}
    \label{fig:overview}
\end{figure*}

\section{RELATED WORKS}

{\em Non-Stationary Reinforcement Learning}---
Very recently, a lot of attention have been drawn to address non-stationary environments through DRL. For instance, \cite{Padakandla2020} proposed online Context-Q learning, based on change point detection method\cite{singh2019change}, that stores various policy distributions from varying contexts occured in known model-change patterns. 
Another work \cite{worstcase_nsmdp} investigated Non-Stationary MDPs (NS-MDP) by considering the non-stationary environment as an adversary to the learning agent and constructed a model based tree search algorithm close to traditional minimax search algorithm. In \cite{xie21c}, authors built probabilistic hierarchical model to present non-stationary Markov chain with sequence of latent variables and used variational inference to set a lower bound for log probability of evidences observed. 

{\em Non-parametric Reinforcement Learning}--- RKHS based non-parametric reinforcement learning and associated rich function class has been used previously to generate stochastic policies. For instance, \cite{pmlr-v38-lever15} proposed compact representation of policy within an RKHS with no need of remetrization of abstract feature space and showed a non-parametric actor-critic framework through efficient sparsification method in RKHS. 
The authors in \cite{SPGT} and \cite{PGT_NSMDP} showed more attention to device a stochastic policy gradient method by plugging unbiased stochastic policy gradient computed in an RKHS and build a theoretical framework of convergence to a neighborhood of critical points. 

In general, recent methods focused to accurately outline control policies in a fixed parameter space while non-parametric methods tried to build a framework for policy representation in an RKHS to exploit its rich function space. In our work, we employed inner-product property of RKHS to build up an adaptive checkpoint detection method for learning optimal control policies in a dynamically changing environment.  
\section{PRELIMINARIES}

We design the problem formulation in the context of non-stationary RL where an agent operates in an environment with unknown non-stationarity, with a motive to learn optimal control policies to maximize the agent's long-term utility. This interaction with non-stationary environment is lately modelled as Non-Stationary Markov Decision Process (NS-MDP) \cite{worstcase_nsmdp}, \cite{chandak20optimforfuture}, \cite{cheung20a}, \cite{Padakandla2020}. Many real-world non-stationary environments have a tractable state-space but it is practically impossible to draw an assumption on how its future evolution will look like \cite{silva2006},\cite{worstcase_nsmdp}. Based on this hypothesis, the agent accumulate snapshots of current model after each action is executed, but it can not easily track the dynamic variance in state distribution. We target to build an adaptive learning framework for handling such dynamic variance. In section III-A, we formulate the layout of NS-MDP framework and in section III-B we explicitly mention the bottlenecks of parametric approach for computing policy integrals with deterministic gradients. In section III-C, we mention {\em non-parametric} policy gradient methods formulated over RKHS.   

\subsection{NS-MDP Composition}
Let $\mathcal{P}(\mathcal{S})$ be a set of probability distribution for each
state $s \in \mathcal{S}$, and the NS-MDP instance be a tuple
$(\mathcal{S,A,T},\{r_t\}_{t\in\mathcal{T}},\{P_t\}_{t\in\mathcal{T}},P_{0})$ \cite{worstcase_nsmdp}, where $\mathcal{S}$ and $\mathcal{A}$ represent
the state and action spaces, $\mathcal{T}:\{1,2....,N\}$ represents each decision epoch when agent is interacting with the environment, $r_t(\cdot \mid a, s) \in \mathcal{P}(\mathcal{R})$
represents state-reward distribution over actions at decision epoch $t$; $P_t(\cdot \mid a, s) \in
\mathcal{P}(\mathcal{S})$ is the state transition probability distribution at decision epoch $t$, which is non-stationary; and
$P_{0}(\cdot)\in\mathcal{P}(\mathcal{S})$ is the probability distribution over initial state $s_{0}$. The agent selects actions using policy
$\pi(\cdot \mid s) \in \mathcal{P}(\mathcal{A})$, which is a probability
distribution over actions, conditioned over state $s$. The sequence of action
selection given a state forms a Markov chain $\xi =
(s_{0},a_{0},s_{1},a_{1},\dots,s_{T-1},a_{T-1},s_{T})$. The probability density function for $\xi$ is represented as
%
$
    \operatorname{Pr}(\xi \mid \pi)=P_{0}(s_{0}) \prod_{i=0}^{T-1} \pi(a_{i} \mid s_{i}) P_t(s_{i+1} \mid s_{i}, a_{i})$ and the discounted reward is defined by
$R(\xi) = \sum_{i=0}^T\gamma^tr_t(s_i,a_i)$ where $\gamma$ is the discount factor. 
\subsection{Bottlenecks in parametric deterministic policy learning}
In the case of conventional stationary MDP setting with discounted reward function, there always exists a Markovian Deterministic stationary action-mapping policy\cite{puterman2014}. 
In case of deterministic policy search algorithm like DDPG, the policy search is done by estimating the gradients over expected return, in form of Q-value function $Q^{\mu}(s,a)$, from of class of parameterized deterministic policies $\mu(\cdot|s,\theta) , s\in\mathcal{S}, \theta\in\Theta$ w.r.t parameter matrix $\theta$. The parameter matrix is adjusted by following the guidance provided by policy gradients computed as $\nabla_\theta J(\mu_\theta) = \int_{\mathcal{S}}Pr(\xi |\mu_\theta)\nabla_\theta \mu_\theta(s)\nabla_aQ^{\mu_\theta}(s,a)|_{a=\mu_\theta(s),s\sim\xi}ds$ \cite{silver2014deterministic}. Firstly, this integration only spans over visited state-space. As a result, a deterministic policy can not model the varying dynamics of a non-stationary environment. Secondly, deterministic estimation of policy gradients are unsound since it assumes that a stationary state distribution and reward function always exist in the MDP setting. This assumption makes the estimated gradients irrelevant if the reward distribution is likely evolving as time passes \cite{BPG}. In such cases, a probabilistic approach such as stochastic policy calculation is more rigid to handle non-stationary transition dynamics. 

In parametric cases, policies are given as linear functions over feature space $\phi(\cdot)$ as $\mu_\theta(\cdot) = \theta^T\phi(\cdot)$ or by non-linear approximators as $\mu_\theta(\cdot) = g(\theta^T\phi(\cdot))$. With a Euclidean Norm, it is very straightforward to compute policy gradients with respect to parameters $\theta$. But, there exists no apparent way to choose a scaling of these parameters. 
Besides, deficient parameter initialization slows down the learning rate and pre-conditioning of the priori becomes computationally expensive (detailed discussion in \cite{pmlr-v38-lever15}).    

\subsection{Non-Parametric Space for Gradient Estimation}

To address the above issues with parametric approaches, we opt to design {\em non-parametric} stochastic policy learning framework with a non-decreasing vector-valued RKHS in this paper. First of all, now policy gradient is an entire function in RKHS, not limited by a fixed priori parameterization of the class. Thus, we can exploit a rich function class for adaptively learning a complex, but sufficient, representation for stochastic policies. 

Here, we opt to represent policies as multivariate
Gaussian models defined as a mapping function
$h:\mathcal{S}\rightarrow\mathcal{A}$, where $h(s)$ belongs to a vector-valued
RKHS $\mathcal{H}$ in which, $\forall h \in \mathcal{H}, \ s \in \mathcal{S} \
\text{and} \ \mathbf{a} \in \mathbb{R}^d$, the reproducing property $\langle
h(\cdot), \mathcal{K}(s,\cdot)\mathbf{a}\rangle_{\mathcal{H}} = h(s)^T\mathbf{a}$,
holds true. Here, $\mathcal{K}(s,s^\prime)$ is a matrix valued Gaussian kernel. Gaussian control policies can be expressed as
\begin{equation}
    \label{eq5:rkhs-pg}
    \pi_{h,\Sigma}(a|s) = \frac{1}{\sqrt{2\pi |\Sigma|}}e^{-\frac{(a-h(s))^T\Sigma^{-1}(a-h(s))}{2}}.
\end{equation}

In this non-parametric learning paradigm, the learning objective will be 
finding the optimal mapping $h^*$ such that
\begin{equation}
    \label{eq6:rkhs-mapping-ftn}
    \scriptsize h^*:= \argmax_{h\in \mathcal{H}}J_{s_0}(h) = \argmax_{h\in \mathcal{H}}\mathbf{E}\Big[\sum_{t = 0}^\infty\gamma^tr(s_t, a_t)|h, s_0\Big].
\end{equation}
Similarly, the action-value function $Q(s,a;h)$ can be defined as
    $Q(s,a;h) = \mathbf{E}[\sum_{t = 0}^\infty\gamma^tr(s_t, a_t)|h, s_0, a_0]$.
Furthermore, by utilizing the formulation of Fr\'echet Derivative as in~\cite{pmlr-v38-lever15}, the
gradient of $J_{s_0}(h,\cdot)$ with respect to functional $h$ yields
\begin{equation}
\label{eq8:}
    \scriptsize \nabla_h J_{s_0}(h,\cdot) =   \frac{1}{1-\gamma} \mathbf{E}_{(s,a)\sim\rho_{s_0}(s,a)}\big[Q(s,a;h)\kappa(s,\cdot)\Sigma^{-1}(a-h(s))|h\big],
\end{equation}
a function in $\mathcal{H}$, where $\rho_{s_0}(s,a) := \pi_h(a|s)(1-\gamma)\Sigma_{t=0}^\infty\gamma^tp(s_t|s_0)=\pi_h(a|s)\rho_{s_0}(s)$. Finally, as the gradient ascent upgrades the mean functional as 
$h_{k+1}=h_k+\alpha_t\nabla_{h_k}J_{s_0}(h_k,\cdot)$,  Eq.
\eqref{eq8:} can be expressed as,
{\scriptsize
\begin{equation}
    \begin{aligned}
         \nabla_h J_{s_0}(h,\cdot) ={}&\frac{1}{1-\gamma}\int\int Q(s,a;h)\kappa(s,\cdot)\Sigma^{-1}(a-h(s))\rho_{s_0}(s,a)dsda\\
        ={}&\frac{1}{1-\gamma}\int D(s) \kappa(s,\cdot)\rho_{s_0}(s)ds, \label{eq9:}
    \end{aligned}
\end{equation}
}
where $D(s) = \int Q(s,a;h)\Sigma^{-1}(a-h(s))\pi_h(a|s)da$ following
\cite{PGT_NSMDP}.
Moreover, modeling probability distributions in an
RKHS (Eq. \eqref{eq9:}) can be principally scaled by RKHS norm, which does not require to retain a parameter vector. We only rely on repositioning (vector-valued version
of) kernels on the support of current snapshots of state-space to predict transition dynamics in evolutionary NS-MDP settings.

\section{PROPOSED METHODOLOGY}
At high level, our proposed \texttt{AdaptiveH} framework is based on Kernel based Value Approximation and {\em non-parametric} policy learning equipped with advantages from kernel based feature mapping within a defined RKHS which adaptively facilitates frequent policy updates when required in a dynamically challenging environment. We extend the theoretical basis described in subsection III-C by doing gradient estimation through kernel based high (possible infinite)-dimensional feature mapping and an inner-product based gradient direction checking for maximizing utility in a highly non-stationary environment. As shown in Fig.\ref{fig:overview}, by only executing the optimal part of the whole trajectory and frequent policy update when required, we promise a sample-efficient learning in RKHS and faster to converge algorithm to handle dynamic variance in the system. 

Considering the simulated agents working in dynamic environments modelled as NS-MDP setting, current policy $\pi_{h_k}(\cdot|s_0)$ suggests an initial action sequence $\mathcal{A} = \{a_0, a_1, ... , a_n\}$ based on current snapshot of MDP $\mathcal{M}_0$. But, since the transition dynamics of the NS-MDP setting and reward function is continually evolving through time, the agent sporadically experiences an urge to upgrade it previous policy $\pi_{h_0}(\cdot|s_0)$ before even fully executing the already {\em in-queue} actions from previous policy. Our principal investigation is centered on this question on how to estimate the exact adaptive time-period window $\texttt{H}$ at which the agent will pre-maturely retreat and upgrade to a new policy to handle hidden non-stationarity of the environment. In \cite{xiao2020thinking}, this time-period $\texttt{H}$ was kept as a fixed value while concurrently execution previous actions and learning new policies.
\begin{figure}[h!]
    \includegraphics[width=\linewidth]{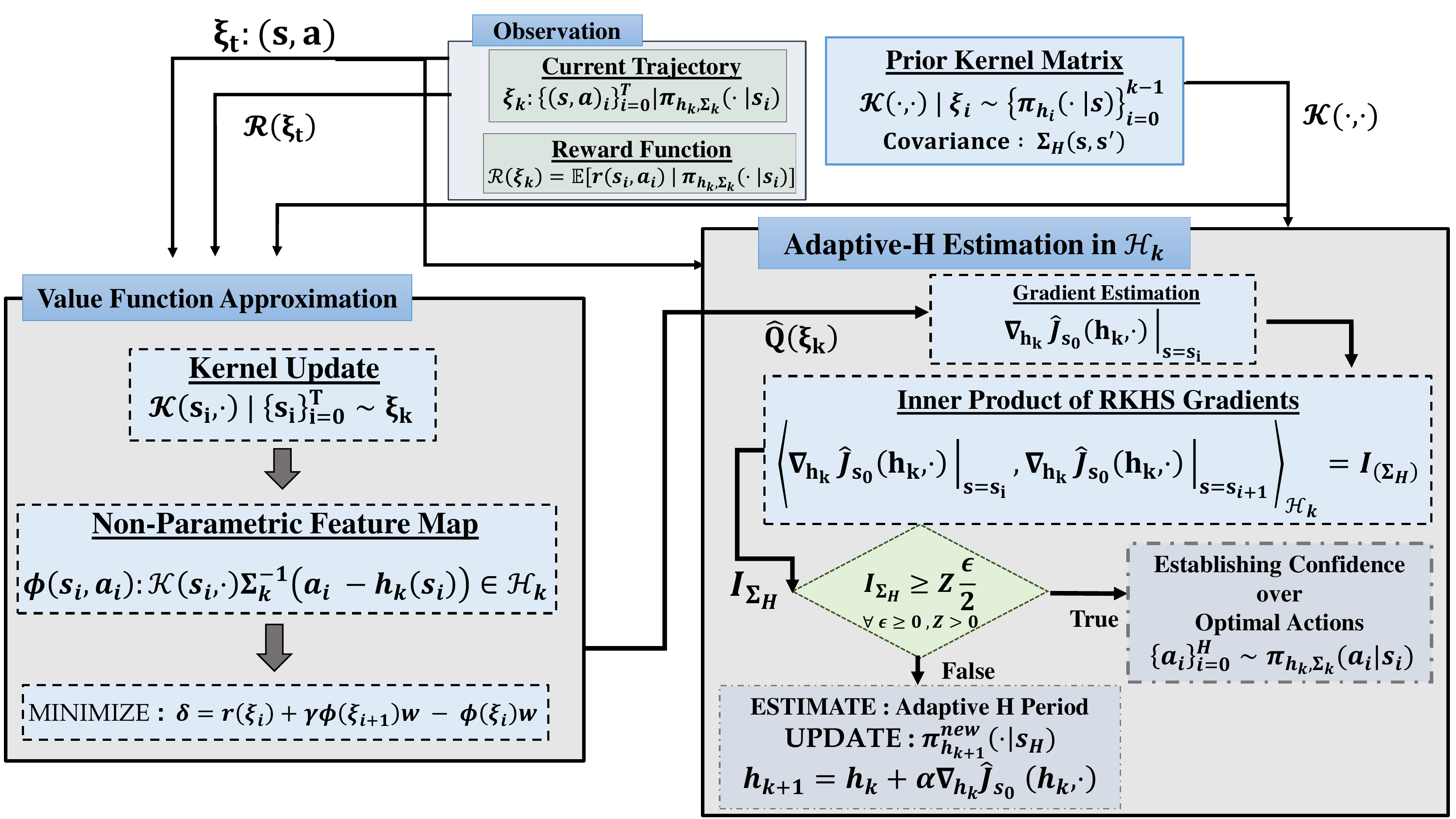}
    \caption{Detailed Algorithmic Approach for \texttt{Adaptive-H} estimation}
    \label{fig:algo}
\end{figure}

Specifically, our approach (in Fig. \ref{fig:algo}) proposes a resolution to challenges of adaptively bisecting the whole sampled $\mathcal{A}$ into optimal $\mathcal{A}_{opt}$ and obsolete $\mathcal{A}_{left}$ parts by using similarity metric in RKHS. Thus, our agent frequently revisit the policy update after only executing those optimal actions compatible to the dynamic changes happened in the environment and our algorithm exhibits sample-efficiency and faster convergence in a NS-MDP setting. In particular to map our state-space into a high (possibly infinite)-dimensional RKHS, we construct diagonal matrix-valued Gaussian kernel, $\kappa_{\Sigma_{H}}(s,\cdot)$ with covariance matrix
$\Sigma_{H}$ as
\begin{equation}
    \label{eq:rbf}
    \scriptsize \kappa_{\Sigma_{H}}(s,s^\prime)_{ii} =
    e^\frac{{-(s-s^\prime)^T\Sigma_{H}^{-1}(s-s^\prime)}}{2} \ \ \ \ \ \ \ \ \ \scriptstyle{\forall i=1,...,m }.
\end{equation}

\subsection{Value Function Regression in Hilbert Space}
Subsequently, for doing Q-value estimation as shown in Fig. \ref{fig:algo}, we utilize kernel based least square temporal difference method \cite{ISinRKHS}, \cite{xu2007kernel} to regress action-value function
$\hat{Q}(s,a)$ and replace the $Q(s,a;h)$ in Eq.
(\ref{eq8:}). With this approximated value, the actor chooses the best action mapping
function $h(s)$ that derives the policy gradient in a close neighborhood of the
optimal point. Following \cite{pmlr-v38-lever15}, given the vector-valued RKHS $\mathcal{H}$, we define a  $\mathcal{H}$ compatible 
feature map $\phi(\xi_k)$ for each new $(s_i,a_i)$ coming from newly observed trajectory $\xi_k$ following $\pi_{h_k}(a_i|s_i)$ as,
$\phi : (s_i,a_i)\rightarrow\mathcal{K}(s_i,\cdot)\Sigma^{-1}(a_i-h_k(s_i)) \in \mathcal{H}_K$.
As shown in Fig.\ref{fig:algo}, we first include new kernel centers $\{s_i\}_{i=0}^T\sim\xi_k$ in prior kernel dictionary and update prior Kernel Matrix to $\mathcal{K}(s_i,\cdot)$. With the updated kernel matrix and feature map $\phi(\xi_k))$, we continually minimize the TD-error \cite{boyan1999least},
$\delta = r(\xi_k) + \gamma\phi(\xi_{k+1})w - \phi(\xi_k)w$
to regress $\hat{Q}(s,a)$ in $\mathcal{H}$.

\subsection{Unbiased Policy Gradient Learning in RKHS}

By estimating the $\hat{Q}(s,a)$ and inputting the values to Eq. \eqref{eq8:}, we can get an unbiased policy gradients for any $(s_i, a_i)$ pair through utilizing compatible kernel $\mathcal{K}(s_i, \cdot)$ defined in same RKHS $\mathcal{H}$. Moreover, since, action-mapping policy $\pi_{h_k}(\cdot|s_i)$ belongs to $\mathcal{H}$, the policy gradient is too embedded over the same $\mathcal{H}$. This simple analytic form of the policy gradient is a blessing popped with embedding policy distributions in a Hilbert Space
This simple yet effective policy gradients are considered as functional gradients in an RKHS and thus gradients are only affected by Hilbert Space norm which ensure smoother control policies in policy space as discussed in section III-C.     
\begin{algorithm}
\SetAlgoLined
\scriptsize

  \SetKwData{Left}{left}
  \SetKwData{Up}{up}
  \SetKwFunction{FindCompress}{FindCompress}
  \SetKwInOut{Input}{input}
  \SetKwInOut{Output}{output}
\textbf{Initialize} 
\\ \quad NS-MDP: $\mathcal{M} = \{\mathcal{S}. \mathcal{A}, r_t, P_t, P_o\}$
\\ \quad Kernel: $\mathcal{K} \in \mathcal{H}:\mathcal{S}\times\mathcal{S}\Rightarrow \mathcal{L}(\mathcal{A})$
\\\quad RBF Covariance: $\Sigma_{H}$
\\\quad Initial Policy: $\pi_{h_o,\Sigma_o}$
\\\textbf{Repeat}
\\ \For{$episode\gets 0, 1, 2 \dots $}{
    $S_{o}\leftarrow  env.reset(rand.seed)$
  \\$s_{o}\leftarrow Initial State Sampling From  \; P_o \,\forall \,s_{o}\in S_{o}$
    \\\For{$i\gets0, 1, 2$ $\dots$}{
    $a_{i} \in \mathcal{A} \gets \pi_{h_k}(\cdot|s_i)$
    \\$s_{i+1},\; r,\; done = env.step(a_{i})$
    {\tiny \# snapshots of NS-MDP $\rightarrow s_{i+1}$}
    \\$\mathcal{K}(s_{i},\cdot)\gets KernelUpdate()$
    \\\textbf{\texttt{COMPUTE}}: 
    $\phi(s_i,a_i)\gets\mathcal{K}(s_i,\cdot)\Sigma^{-1}(a_i-h_k(s_i))$\\
    \ \ \ \ \ \ \ \ \ \ \ \ \  $\hat{Q}_{\pi_{h_k}}(s_i,a_i) \gets kernelQApproximation()$\\
    \ \ \ \ \ \ \ \ \ \ \ \ \  $\scriptstyle\nabla_{h_{k}}\hat{J}_{s_o}(h_k,\cdot) \gets \hat{Q}(s_i,a_i)\mathcal{K}(s_i,\cdot)\Sigma^{-1}(a_i-h_{k}(s_i))$ 
    \\\textbf{\texttt{EVAL}}: $\scriptstyle\langle\nabla_{h_{k}}\hat{J}_{s_o}(h_k,\cdot)|_{s_o},\nabla_{h_{k}}\hat{J}_{s_o}(h_k,\cdot)|_{s_i}\rangle \geq \texttt{threshold}$ \\
    \If{\textbf{FALSE}}
    {\textbf{\texttt{UPDATE}}: \\ \quad Gradient Ascent Step : $\scriptstyle h_{k+1} \gets h_k + \alpha \nabla_{h_k}J_{s_o}(h_k,\cdot)$
    \\ \quad Upgraded Policy : $\scriptstyle\pi_{h_{new}}:\pi_{h_k}(\cdot|s_i) \gets \pi_{h_{k+1}}(\cdot|s_i)$ 
    }
    \textbf{\texttt{FIT}}: Minimize TD Error: $\scriptstyle\delta = r + \gamma\phi(s_{i+1},a_{i+1})w - \phi(s_i, a_i)w$ \\
    \texttt{SPARSIFY} : Kernel Matrix $\mathcal{K}(\cdot,\cdot)$
    
    }}
 \caption{Non-Parametric Adaptive Learning}
\end{algorithm}

\begin{figure*}[t]
    \includegraphics[width=1\textwidth]{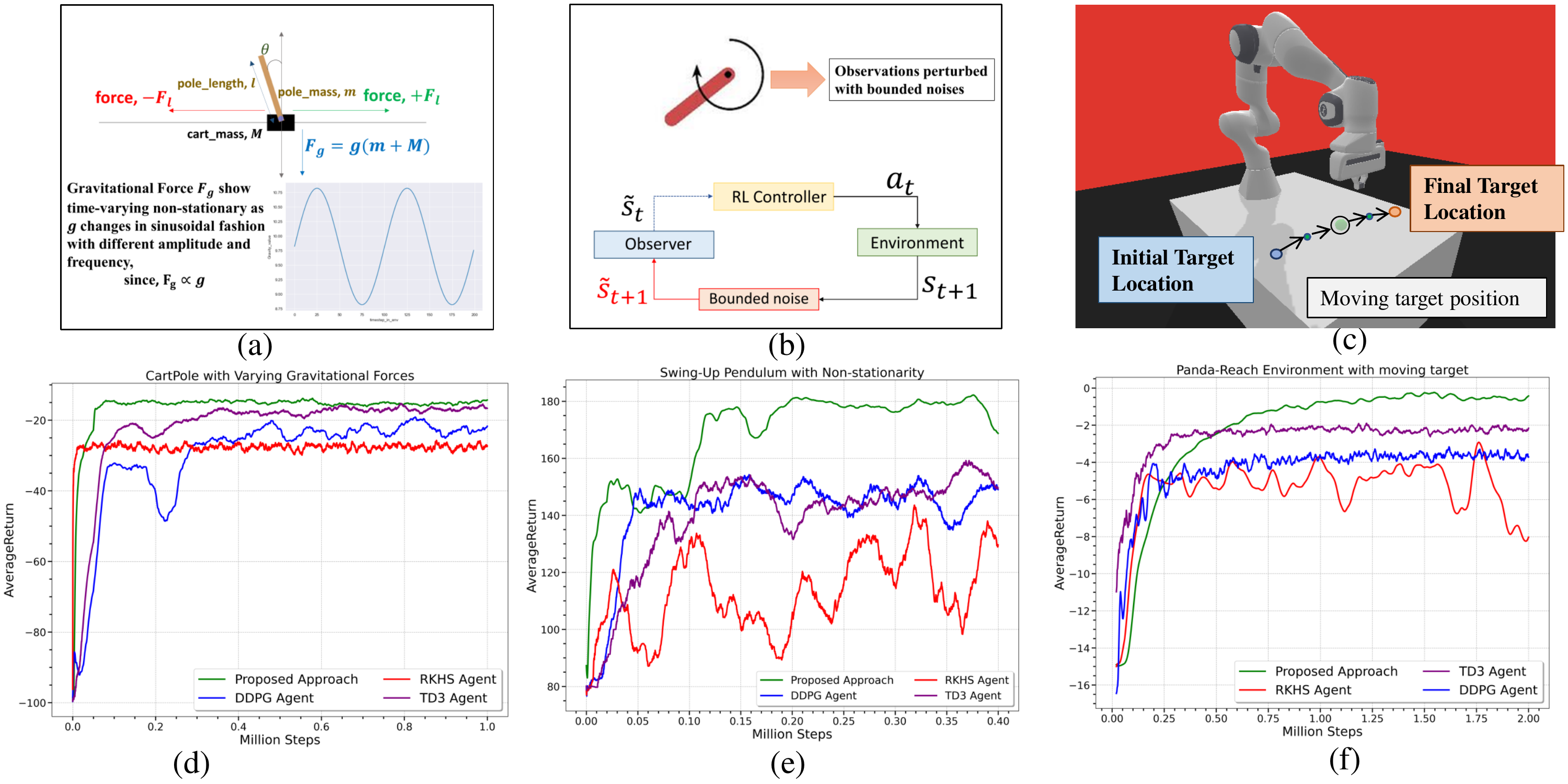}
    \caption{(a) NS Continuous CartPole environment, (b) NS Swing-up Pendulum Environment, (c) NS Reach Task, (d)-(e)-(f) learning curve comparison between proposed approach and baselines}
    \label{fig:main_results}
\end{figure*}

\subsection{Inner-product based Evaluation for \texttt{Adaptive-H}}
RKHS possesses all properties of a Vector Space and we can compute the inner-product between consecutive policy gradients estimated through any trajectory $\xi_k$ sampled by current policy $\pi_{h_k}(\cdot|s_i)$. The kernel matrix $\mathcal{K}(\cdot, \cdot)$ encodes the prior experiences with the covariance shift happened for underlying non-deterministic transition dynamics and $\mathcal{K}(\cdot, \cdot)$ can be regarded as a measure of proximity between different kernels in Hilbert Space in turn defining complex function space within which we can detect sequence of control policies for handling dynamic variance. This {\em non-parametric} representation of policy distributions grants an adaptive remolding of the old policy $\pi_{h_k}$ into more upgraded new policy for continual state-space evolution. 

The principal essence of this work is the introduction of an inner-product based similarity checkpoint between gradient updates for setting adaptive period defined as $\texttt{H}^\prime$ in Fig.\ref{fig:overview}. For reward maximization in RL setting, the gradients must follow an ascent direction in the solution space. By Eq. \eqref{eq8:}, policy gradients $\nabla_{h_k}J_{s_0}(h_k,\cdot)|_{s_i}$ with respect to current policy mean functional $h_k$ for current trajectory $\xi_k$ can be easily measured. With consecutive policy gradient functionals, an inner product $\langle\nabla_{h_k}J_{s_0}(h_k,\cdot)|_{s_0},\nabla_{h_k}J_{s_0}(h_k,\cdot)|_{s_{i}}\rangle_{\mathcal{H}}$ can be calculated as both policy gradients belong to same RKHS. This inner product based similarity evaluation adjusts the confidence on current actions and adaptively sets the time-frame window $\texttt{H}^\prime$. Besides, after each optimal action execution, current snapshot of NS-MDP is recorded for gradient estimation. Since, the state transition function and reward distribution is continually evolving, parts of previously suggested actions $\{a_i\}_{i=0}^{n}$ from previous mapping $h_k$ will become sub-optimal and there emerges an appeal to relearn new mapping $h_{k+1}$ in between a full episodic interaction. Our proposed approach suggests that optimal point to prematurely upgrade to new policy. Theoretically, with respect to the underlying non-stationarity, $h_k$ will be outside of an $\epsilon$ neighborhood of the optimal solution i.e $|\nabla_{h_k}J_{s_0}(h_k)| > \epsilon$. We need to update our policy and learn new action mapping functional $h_{k+1}$ whenever the policy gradients do not follow an ascent direction during optimization. Whenever, the condition $\langle\nabla_{h_k}J_{s_0}(h_k,\cdot)|_{s_0},\nabla_{h_k}J_{s_0}(h_k,\cdot)|_{s_{i}}\rangle_{\mathcal{H}} \geq Z\epsilon/2$ violates during finishing actions from old policy, the agent immediately retreats and do a revisit to policy upgrade for more updated actions. 
Mathematical proof of this statement is (inspired by \cite{PGT_NSMDP}) shown as Theorem 1 in Appendix. We added the Appendix in \color{blue}\url{https://sites.google.com/view/adaptiveh}\color{black}. We have discussed thoroughly and elaborately the mathematical validations of our proposed method in terms of RKHS. Moreover, this positive lower bound on the dot product based evaluation metric also turns the dynamic regret term $R_k$ into non-negative sub-martingale, assuring convergence of our proposed framework as $R_k:=J_{s_0}(h_k^*)-J_{s_0}(h_k)$--- converges to its optimal value as $\mathbb{E}[R^*] = \lim_{k\rightarrow\infty}R_k$. This is defined as well in detail in Theorem 2 of Appendix. Thus, our proposed approach confirms a fixed-point solution for the non-stationary environments.

\section{Experimentation Platforms}

Through experiments, we focused to validate our central finding that existing state-of-the-art parametric method, such as TD3 and DDPG struggle to contrive a perfect model of the environment perturbed with persistent non-stationarity, while stochastic policy distribution represented non-parametrically in a vector-valued RKHS can emerge as an efficient substitute for various benchmarks. Moreover, by launching frequent, but adaptive upgrades from current policy to new policy, our learning algorithm has ensured faster learning convergence and sample-efficiency.

{\bf Environments--}We replicated non-stationarity by molding two classic control benchmark problem with continuous perturbation of physics parameters or by coupling observations with noises. Besides, we have also used a high dimensional simulated robotics environment based on Panda-Gym \cite{gallouedec2021pandagym} where a 7 Degree of Freedoms (DoFs) Franka Emika Panda Arm has to execute optimal actions to reach continuously translating target position smoothly and swiftly.
\textit{NS CartPole---}The first environment is the classic problem called Cartpole where the agent need to balance a pole on a moving cart by applying torques on the cart. We constructed a continuous action-space variant of this environment based on the discrete one from OpenAI Gym\cite{brockman2016openai}. We consider changes by continuously varying the magnitude, within bounds, of gravitational force, a physics parameter of the system as drawn in Fig.\ref{fig:main_results}(a), and the agent needs to dynamically adapt to this variation. \textit{NS InvertedPendulum---}The second environment is the Swing-Up Pendulum environment. In this problem, we cluttered observation-space with bounded random noises to reproduce real-world latency and noise as shown in Fig.\ref{fig:main_results}(b). \textit{NS ReachTask---}The last environment is based on a high-dimensional robotics environment where we introduced a real-world non-stationarity where the target location for a 7 DoFs robotic manipulator is continuously moving through a varying, but bounded, speed at a particular direction as shown in Fig.\ref{fig:main_results} (c). Here, we assured that the target location always remains within the reachable workspace of the Panda Arm.      

\begin{figure}[htbp]
    \includegraphics[width=\linewidth]{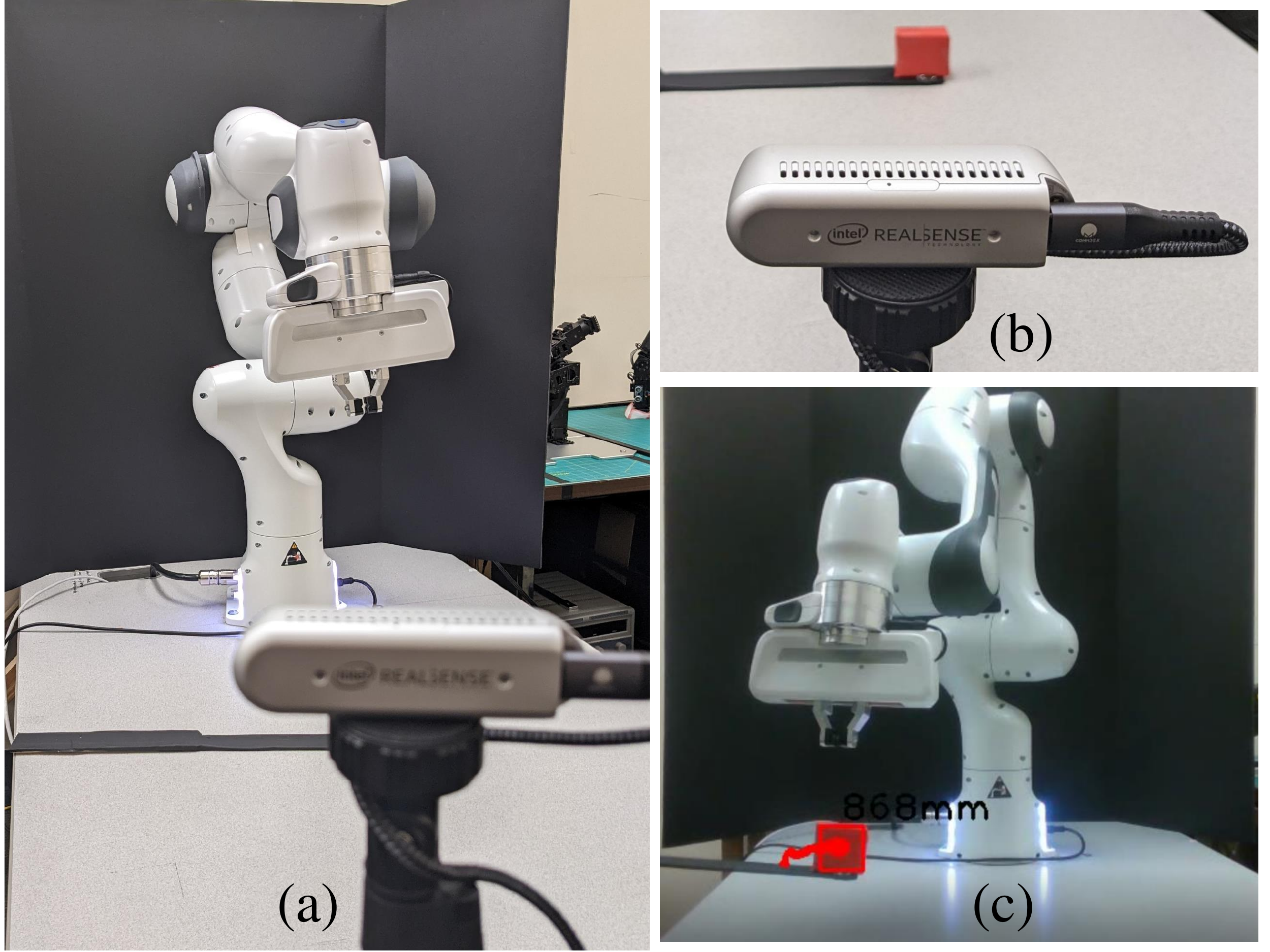}
    \caption{(a) Hardware setup with a 7 DOFs Franka Emika Panda Arm (b) Intel RealSense Depth Sensing Camera as tracking sensor to track the position of the moving target position(red colored box), (c) Position tracking through Depth Sensing Camera.}
    \label{fig:Hard}
\end{figure}

\textbf{Real Hardware Setup--} 
To validate our study in a real world scenario, we have mostly focused on the simulated robotic non-stationary environment to perform the real-hardware experimentation and we have used here the same 7-DoF Franka Emika Panda robot arm, that is mounted on
table-top similar to simulation environment as shown in Fig.\ref{fig:Hard}(a). Besides, we
have utilized here a Intel RealSense Depth Camera D435i
for tracking the dynamically moving target location and extracting the depth
information to obtain the current target position in 3D coordinate frame. This information is fed to RL controller as current 3D target location is a part of the state-space vector for our RKHS based learning algorithm. The Franka Arm integrated library--libfranka and Franka ROS turns the communication protocol between simulation and hardware very trouble-free through low-latency and low-noise communication based on ROS ecosystem.

\section{Results Analysis}
Our experiments seek to explore the following queries:
\begin{enumerate}
    \item Does introducing kernel-centric stochastic policy learning in an RKHS deliberately showcase better performance than baselines in a non-stationary settings? 
    
    \item How much does the degree of non-stationary inference affect the performance of our \texttt{AdaptiveH} approach?
    
    \item Does the proposed infinite-dimensional feature space initiate sequence of optimal policies at perfect interruption point for handling varying dynamics?
\end{enumerate}

\textit{Better Learning Performance in Non-Stationary settings--}
The baselines to validate our proposed methodology are TD3 and DDPG implementation from OpenAI Baselines\cite{stable-baselines3} and RKHS non-parametric policy gradient. Our evaluation criteria is the cumulative average return the agent gains on each trial. 
Learning curves are plotted in Fig. \ref{fig:main_results} (d)-(f) for three respective environments. From the learning curves drawn for environments, it is apparent that our proposed method promisingly outperformed all the baselines. Our method largely stabilizes the learning curves as episodic interaction increases, where the baseline DDPG exhibits relatively high variance in both \textit{NS CartPole} and \textit{NS InvertedPendulum} environments. Baseline agent--TD3 shows relatively better performance but it takes larger amount of training time to reach convergence in \textit{NS CartPole} and in case of \textit{NS InvertedPendulum}, TD3 agent lacks any promising performance due to uncontrolled stochastic perturbations. Moreover, the baseline RKHS method traps inside local {\em minima}, since it can not accomplish frequent policy update when required. Our method grants faster convergence and better sample-efficiency. 
In \textit{NS ReachTask}, from the learning curve shown in Fig.\ref{fig:main_results}(f), it is apparent that our proposed approach has earned the maximum utility, while the baseline methods TD3 and DDPG stabilize but can not guarantee of achieving maximum rewards. Since, this is a high-dimensional environment, kernel based method required more training time to accumulate adequate amounts of experiences to build the kernel matrix and handle the dynamic shift through optimal action execution. 
\begin{figure}[htbp]
    \includegraphics[width=\linewidth]{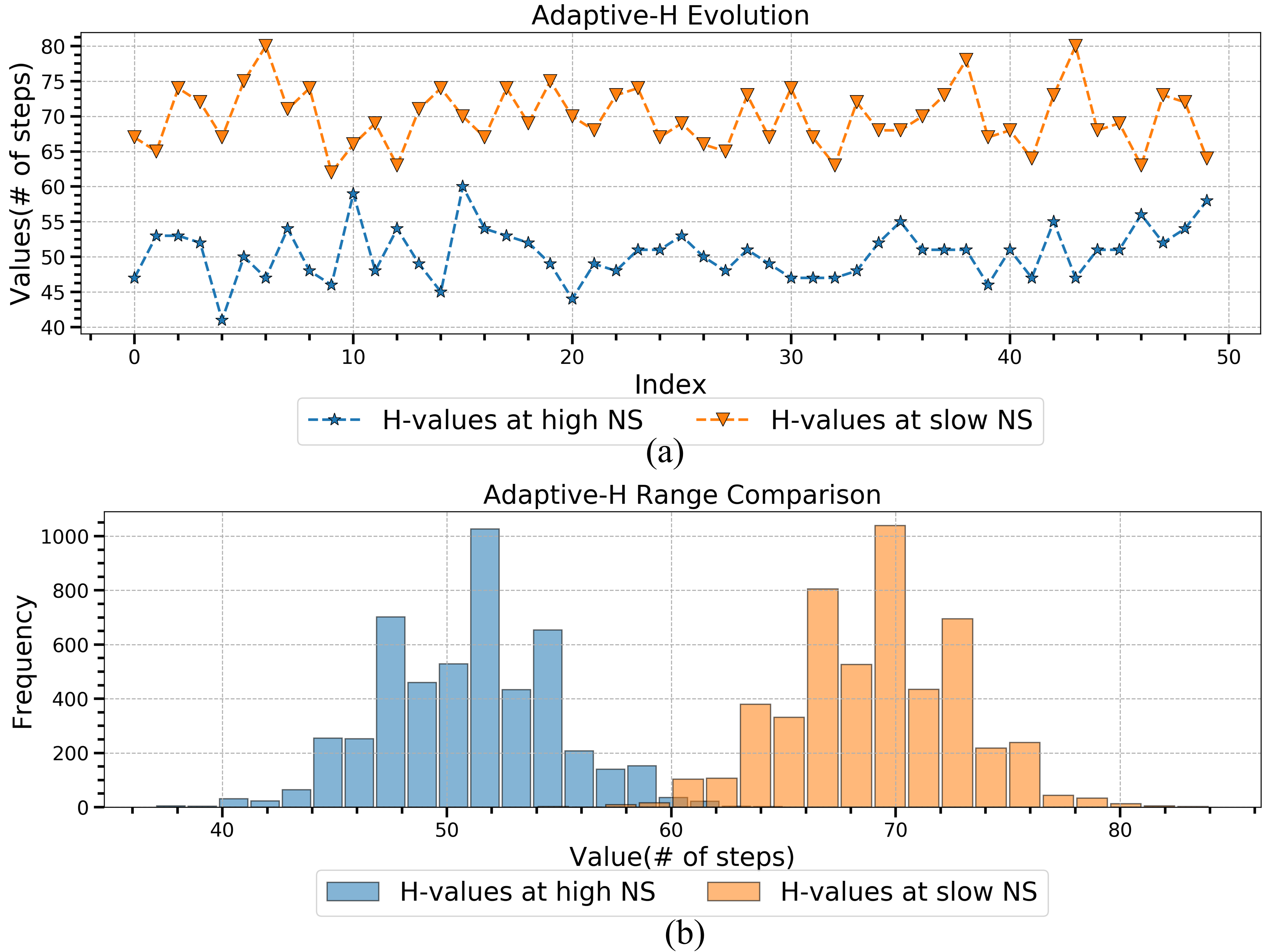}
    \caption{(a) Evolution of window-\texttt{H} of optimal actions for in Cart-Pole environment. (b) Comparison to differentiate range of \texttt{H}-values at two different non-stationary level.}
    \label{fig:H}
\end{figure}

\begin{figure*}[htbp]
    \includegraphics[width=1\textwidth]{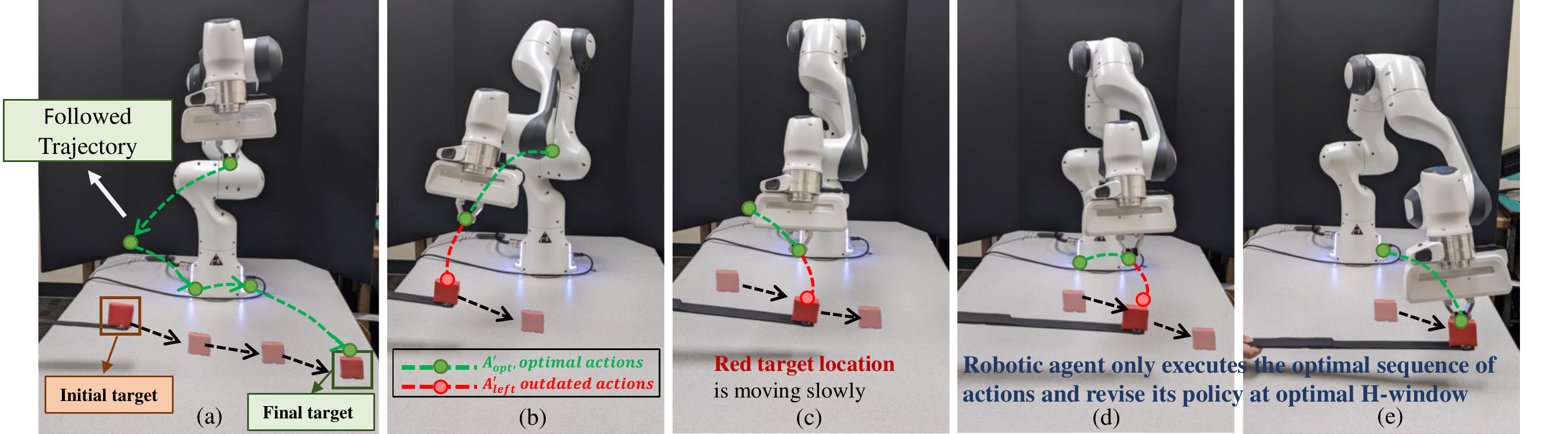}
    \caption{(a) Followed Adaptive Trajectory and dynamically varying target location, (b) only optimal part of action sequence executed and robot upgrades its policy at interruption point, (c)-(d) robot adaptively replans its trajectory to handle moving dynamics, (f) robotic agent swiftly reaching the target location}
    \label{fig:res1}
\end{figure*}

{\em Optimal Action Executions within \texttt{AdaptiveH} Window---} Our proposed approach enables the learning agent to bisect the initial action sequence into two parts-optimal actions and outdated actions. 
The algorithm suggests the agent to upgrade its current action-mapping functional at the action interruption point and the updated policy samples more compatible actions to address the varying dynamics of the environment. We have experimented it in real-hardware setup using the Panda Arm. We have added sequence of snapshots of adaptive policy adjustment and action execution in Fig.\ref{fig:res1}. We showcased that our robotic entity only executes the optimal part of action sequences to swiftly reach its dynamic target location and revisit the policy upgrade several times to handle the varying dynamics of moving target location.    

{\em Varying Degree of Non-stationarity---} We also evaluated our method in \textit{NS CartPole} with more high-frequency gravitational changes to authenticate whether the proposed approach can generalize to faster environmental shift. 
In Fig.\ref{fig:H}, we drew a comparison on how the adaptive window \texttt{H} varies with different magnitude of non-stationary shifts. As shown in Fig.\ref{fig:H}(a), when the environment possesses higher variance, then \texttt{H} values narrow down to lower values. Also, In Fig.\ref{fig:H}(b), it is very clear that for slower non-stationary benchmarks, periods of optimal actions are higher. Reasonably, when the transition dynamics is shifting very fast, but within bounds, policy upgrade is also required to be more frequent to maximize agent's performance. 

\begin{figure}[htbp]
    \includegraphics[width=\linewidth]{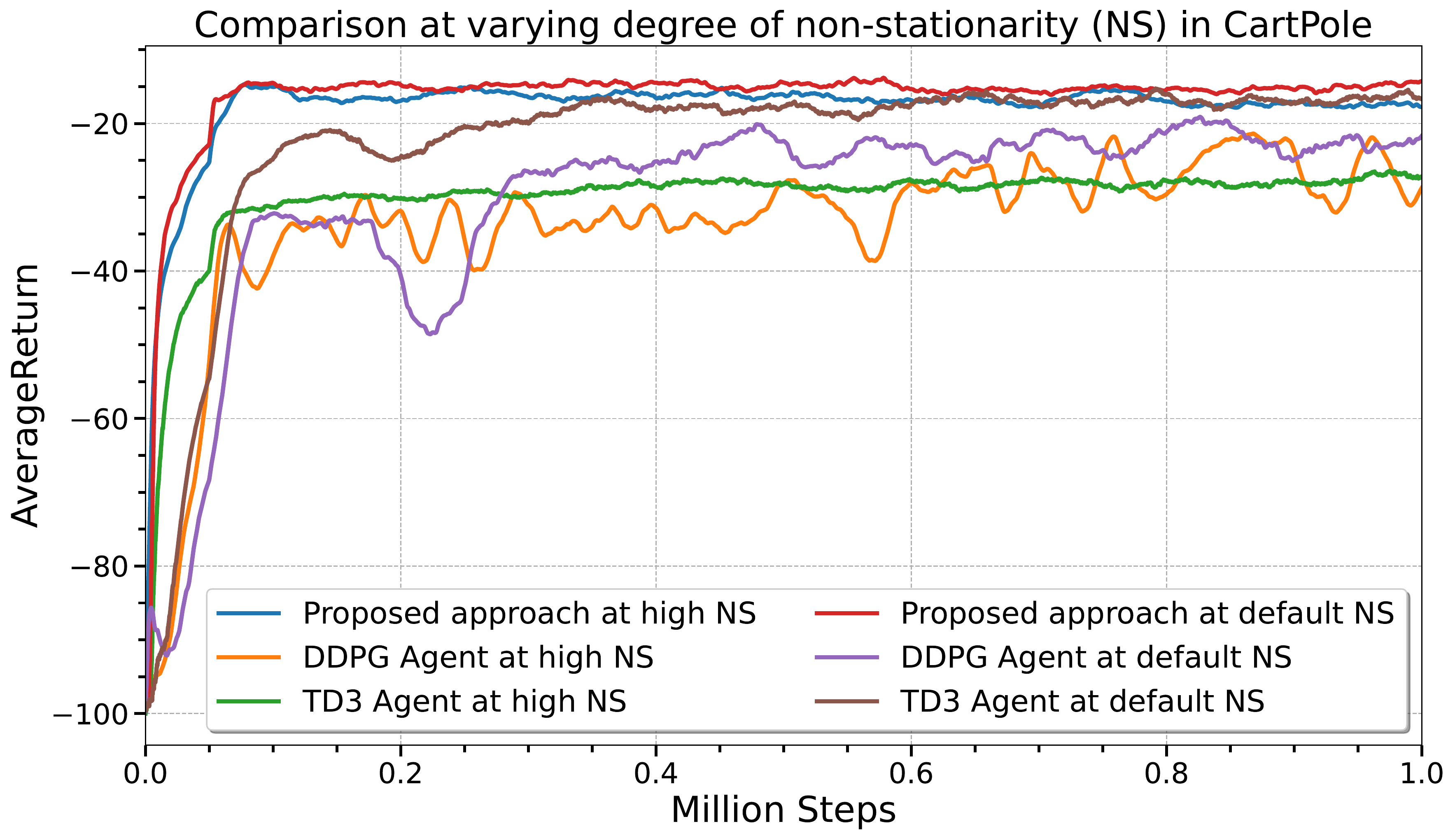}
    \caption{Comparison when degree of non-stationary is changed}
    \label{fig:fast_comparison}
\end{figure}

The learning curve for different non-stationarity has been shown in Fig. \ref{fig:fast_comparison}. From the figure, it is remarkably vivid that in a more complex non-stationary environment, our agent can function more optimally than both baselines. 
Although, the performance for the proposed approach while dealing with high stochasticity slightly worsens, it is totally interpretable that learning control-policies will be strategically strenuous considering high complexity. Besides, the baseline TD3 agent performs admirably better when the environment is perturbed with slower dynamic alterations. But still our proposed approach achieves faster convergence since the agent is empowered with additional dynamic adaptable ability in accordance to system variance.      

\section{Conclusion}
Our adaptive non-parametric learning framework,
through exploiting the inner-product
similarity evaluation in RKHS, 
has outperformed the classic TD3 and DDPG for multiple standard  
benchmarks modified with dynamically varying transition probabilities and
reward functions. 

Maybe most importantly, 
our results have clearly demonstrated,  
both theoretically and empirically, 
the significant benefits of estimating self-adaptive window \texttt{H} 
in generating the sequence of optimal control
policies that maximizes the utility values.
In future
research, we plan to investigate more complex
robotic scenarios in dynamically challenging
environments such as highly cluttered work space and stochastic sensor modality.

\bibliographystyle{IEEEtran}

\bibliography{IEEEexample}



\section*{Supplementary material}

# Supplementary Materials

Apan Dastider and Mingjie Lin

## I. Preliminaries

In this work, the main motivation is to solve concurrently evolving environment by finding sequence of policies $\{\pi_0, ...., \pi_\infty\}$ that maximize the expected discounted cumulative reward of an agent. Let, $\mathcal{S} \subset \mathbb{R}^n$ be a compact set denoting the state space of the agent and $\mathcal{A} \subset \mathbb{R}^d$ be the action space. The transition dynamics are governed by conditional probability $p(s_{t+1}|s_t, a_t \in \mathcal{S} \times \mathcal{A})$ which is non-stationary. We adopt to represent policies as multivariate Gaussian models represented as a mapping function $h : \mathcal{S} \to \mathcal{A}$, where $h(s)$ belongs to a vector-valued RKHS $\mathcal{H}$ in which $\forall h \in \mathcal{H}$, $s \in \mathcal{S}$ and $\mathbf{a} \in \mathbb{R}^d$ the reproducing property $\langle h(\cdot), \kappa(s, \cdot)\mathbf{a} \rangle_\mathcal{H} = h(s)^T \mathbf{a}$, holds true. Here, $\kappa(s, \hat{s})$ is a matrix valued Gaussian kernel with covariance matrix $\Sigma_h$. The policy of the agent can be defined as:

$$\pi_{h,\Sigma}(a|s) = \frac{1}{\sqrt{2\pi det(\Sigma)}} e^{-\frac{(a-h(s))^T \Sigma^{-1}(a-h(s))}{2}}$$

In this non-parametric learning paradigm, the objective is to find sequence of optimal mapping $\{h_0^*, ..., h_\infty^*\}$ such that for each $h^*$

$$h^* := \underset{h \in \mathcal{H}}{\operatorname{argmax}}\, J_{s_0}(h) = \underset{h \in \mathcal{H}}{\operatorname{argmax}} \mathbf{E}\Big[\sum_{t=0}^\infty \gamma^t r(s_t, a_t) | h, s_0\Big]$$

The action-value function $Q(s, a; h)$ : for each state action pair $(s, a)$ produced by mapping $h$ is defined by,

$$Q(s, a; h) = \underset{h \in \mathcal{H}}{\operatorname{argmax}} \mathbf{E}\Big[\sum_{t=0}^\infty \gamma^t r(s_t, a_t) | h, s_0, a_0\Big]$$

By utilizing the formulation of Frechet Derivative [1], the gradient of $J_{s_0}(h, \cdot)$ with respect to functional $h$ yields,

$$\nabla_h J_{s_0}(h, \cdot) = \frac{1}{1-\gamma} \mathbf{E}_{(s,a) \sim \rho_{s_0}(s,a)} \big[Q(s,a;h) \kappa(s,\cdot) \Sigma^{-1}(a - h(s))\big]$$

producing a function in $\mathcal{H}$, where discounted long-run probability distribution $\rho_{s_0}(s, a) := \pi_h(a|s)(1-\gamma)\Sigma_{t=0}^\infty \gamma^t p(s_t|s_0) = \pi_h(a|s)\rho_{s_0}(s)$. Thus, we divide the distribution into two parts: action mapping distribution and state distribution. The distribution of the MDP under $h$, $p(s_t = s, a_t = a|s_0) := \pi_h(a_t|s_t) \int \prod_{u=0}^{t-1} p(s_{u+1}|s_u, a_u) \pi_h(a_u|s_u) d\mathbf{a}_{t-1} d\mathbf{s}_{t-1}$. Then the expectation based gradient equation mentioned above can be expressed as the following integral;

$$\nabla_h J_{s_0}(h, \cdot) = \frac{1}{1-\gamma} \int \int Q(s,a;h) \kappa(s,\cdot) \Sigma^{-1}(a - h(s)) \rho_{s_0}(s,a) ds da$$
$$= \frac{1}{1-\gamma} \int D(s) \kappa(s,\cdot) \rho_{s_0}(s) ds$$

where, $D(s)$ simplifies the term by encapsulating all action based terms as $D(s) = \int Q(s,a;h) \Sigma^{-1}(a - h(s)) da$. Finally, the ascent gradient algorithm improves $h_{k+1} = h_k + \alpha_t \nabla_h J_{s_0}(h_k, \cdot)$.

## Appendix

**Assumption 1:** There exists $B_r$ such that $\forall (s,a) \in \mathcal{S} \times \mathcal{A}$, the reward function $|r(s,a)| \le B_r$. In addition, $|r(s,a)|$ has bounded first and second derivatives with respect to state and action with bounds $B_{rs}$ and $B_{ra}$, $|\partial r(s,a)/\partial s| \le B_{rs}$ and $|\partial r(s,a)/\partial a| \le B_{ra}$

**Assumption 2:** The probability distribution $\rho_{s_0}(s)$ is bounded away and bounded above from zero with constants $B_1 > 0$ and $B_2 > 0\ \forall\ s \in \mathcal{S}$ and $\forall\ h \in \mathcal{H}$ i.e $B_1 \ge \rho_{s_0}(s) \ge B_2$. Also, the transition probability $p(s_{t+1}|s_t.a_t)$ maintains Lipschitz continuity properties with Lipschitz Constant $L_p$, i.e.$|p(s_{t+1}|s_t.a_t) - |p(s'_{t+1}|s_t.a_t)| \le L_p |s_{t+1} - s'_{t+1}|$.

The smoothness properties of the probability distribution is also Lipschitz function with constant $L_{ps}$ and $L_{pa}$, i.e.,

$$|p'(s,a) - p'(s',a')| \le L_{ps} ||s - s'|| + L_{pa} ||a - a'||$$
$$\text{where, } p'(s,a) = \frac{\partial p(s_{t+1}|s_t.a_t)}{\partial a_t} |_{s_t=s, a_t=a}$$

**Lemma 1:** $K_{\Sigma_H}(s, s')_{ii}$ defines a matrix-valued Gaussian Kernel with covariance matrix $\Sigma_H$ such that for $i = 0, 1..., m$, we have that

$$K_{\Sigma_H}(s, s')_{ii} = e^{-(s-s')^T \Sigma_H^{-1}(s-s')/2}$$

When $i \ne j$, then $K_{\Sigma_H}(s, s')_{ij} = 0$. Along with all constants defined in assumption 1 and assumption 2, we further define some constants,

$$B_D = \frac{\sqrt{2} B_r}{1-\gamma} \frac{\Gamma(\frac{p+1}{2})}{\Gamma(\frac{p}{2})} \text{ where } \Gamma(.) \text{ is a gamma function.}$$
$$L_{Qs} = B_{rs} + \frac{\gamma B_r}{1-\gamma} L_{ps} |\mathcal{S}|$$

$$L_{Qa} = B_{ra} + \frac{\gamma B_r}{1-\gamma} L_{pa}|\mathcal{S}|$$
$$L_h = ||h||\lambda_{min}\Gamma_H^{-1/2}$$
$$L_D = L_{Qs} + L_{Qa}L_h$$

Then. we have that $D(s)\rho_{s_0}(s)$ is bounded by $B$ where $B := B_2 B_D$ and maintains all properties of Lipschitz Continuous function with constant $L := B_D L_p + B_2 L_D$.

*Proof:* This proof is very similar to the proof of [[2], Lemma 2].

The proof of Lemma 2 will be divided into two parts.

- Under the statements of assumption 2, we will state that the state probability distribution $\rho_{s_0}(s)$ with initial state at $s_0$ will be a Lipschitz function with Lipschitz constant $L_p$, where $L_p$ possesses same definition mentioned in Assumption 2.
- The term D(s) will be bounded by $B_D$ and Lipschitz with constant $L_D$, where $B_D$ and $L_D$ have the definition as defined in Lemma 2.

Let's start with proof of $\rho_{s_0}(s)$ being a Lipschitz function.

In the preliminaries, $\rho_{s_0}(s)$ is defined by,

$$\rho_{s_0}(s) = (1-\gamma)\sum_{t=0}^{\infty}\gamma^t p(s_t = s|s_0) \tag{1}$$

By marginalization, the term $p(s_t = s|s_0)$ in right hand side of above equation can be expanded as,

$$p(s_t = s|s_0) = \int p(s_t = s, a_{t-1}, s_{t-1}|s_0)ds_{t-1}da_{t-1} \tag{2}$$

Applying Bayes' Rule and Markov property of transition probability of an MDP, Eq. (2) can be further defined as,

$$p(s_t = s|s_0) = \int p(s_t = s|s_{t-1}, a_{t-1})\pi_h(a_{t-1}, s_{t-1})p(s_{t-1}|s_0)ds_{t-1}da_{t-1} \tag{3}$$

Assumption 2 states that transition probability $p(s_{t+1}|s_t, a_t)$ is Lipschitz continuous and bounded with constant $L_p$. Thus using Assumption 2, we can upper bound $|p(s_t = s|s_0) - p(s_t = s'|s_0)|$ by following expression,

$$|p(s_t = s|s_0) - p(s_t = s'|s_0)| = \Big|\int p(s_t = s|s_{t-1}, a_{t-1})\pi_h(a_{t-1}, s_{t-1})p(s_{t-1}|s_0)ds_{t-1}da_{t-1} -$$
$$\int p(s_t = s'|s_{t-1}, a_{t-1})\pi_h(a_{t-1}, s_{t-1})p(s_{t-1}|s_0)ds_{t-1}da_{t-1}\Big|$$
$$= |p(s_t = s|s_{t-1}, a_{t-1}) - p(s_t = s'|s_{t-1}, a_{t-1})|\int \pi_h(a_{t-1}, s_{t-1})p(s_{t-1}|s_0)ds_{t-1}da_{t-1}$$
$$\leq L_p||s - s'||\int \pi_h(a_{t-1}, s_{t-1})p(s_{t-1}|s_0)ds_{t-1}da_{t-1}$$

Since, both probability distributions $\pi_h(a_{t-1}|s_{t-1})$ and $p(s_{t-1}|s_0)$ integrate to one, above expression can be simplified as $|p(s_t = s|s_0) - p(s_t = s'|s_0)| \leq L_p||s-s'||$. Now by using Eq. (1), we can write that,

$$|\rho_0(s) - \rho_0(s')| = (1-\gamma)|\sum_{t=0}^{\infty}\gamma^t(p(s_t = s|s_0) - p(s_t = s'|s_0)|$$
$$\leq (1-\gamma)\sum_{t=0}^{\infty}\gamma^t|(p(s_t = s|s_0) - p(s_t = s'|s_0)| \quad \text{[Using triangular inequality]}$$
$$= \gamma|(p(s_t = s|s_0) - p(s_t = s'|s_0)| \tag{4}$$

The Eq. (4) can be further bounded by using the bound of $|p(s_t = s|s_0) - p(s_t = s'|s_0)|$ as,

$$|\rho_0(s) - \rho_0(s')| \leq L_p||s - s'|| \tag{5}$$

This concludes the proof of part 1. Now we need to proof that action related term D(s) is also Lispschitz continuous and bounded function. From the preliminaries section, $D(s) = \int Q(s,a;h)\Sigma^{-1}(a-h(s))\pi_h(a|s)da$. If we introduce the change of variable $\zeta = \Sigma^{-1/2}(a-h(s))$ and change all variables in the expression of $D(s)$, we have that, $D(s) = \int Q(s, h(s) + \Sigma^{1/2}\zeta)\frac{\zeta}{\sqrt{2\pi\Sigma}}e^{-||\zeta||^2/2}d\zeta$. where, $a$ in original expression of $D(s)$ is replaced as,

$$\zeta = \Sigma^{-1/2}(a - h(s))$$
$$\text{or, } \zeta\Sigma^{1/2} = (a - h(s))$$
$$\text{or,} a = h(s) + \zeta\Sigma^{1/2}$$

and $d\zeta = \Sigma^{-1/2}da$. Moreover, action mapping policy $\pi_h(a|s)$ is defined by $\pi_h(a|s) = \frac{1}{\sqrt{2\pi\Sigma}}e^{-(a-h(s))^T\Sigma^{-1}(a-h(s))/2} = \frac{1}{\sqrt{2\pi\Sigma}}e^{-||\zeta||^2/2}$. To simplify the new expression of $D(s)$, we can define $\frac{1}{\sqrt{2\pi}}e^{-||\zeta||^2/2} = \phi(\zeta)$. So, $\nabla\phi(\zeta) = -2\zeta\phi(\zeta)/2 = -\zeta\phi(\zeta)$. So, previous expression of $D(s)$ can be simplified as,

$$D(s) = -\int Q(s, h(s) + \Sigma^{1/2}\zeta)\frac{\nabla\phi(\zeta)}{\Sigma^{1/2}}d\zeta \tag{6}$$

The above formulation makes the following derivations understandable. Now, the formulation of Integration by parts is given by, $\int uv dx = u\int v dx - \int u'(\int v dx)dx$, where $u$ and $v$ are both functions of $x$. If we apply the formulation of integration by parts in Eq. (6), we have for each $i = 1, ..., n$ that

$$D(s)_i = \int \frac{1}{\Sigma^{1/2}} Q(s, h(s) + \Sigma^{1/2}\zeta)\phi(\zeta)|_{\zeta_i=-\infty}^{\zeta_i=\infty} d\bar{\zeta}_i + \int \frac{1}{\Sigma^{1/2}} \frac{\delta Q(s, h(s) + \Sigma^{1/2}\zeta)}{\delta \zeta_i}\phi(\zeta)d\zeta \quad (7)$$

where, $\bar{\zeta}_i$ denotes the integral with respect to all variables in $\zeta$ except for $i^{th}$ component. By the bound over $Q(s,a)$ function introduced in (44) and since, $\phi(\zeta)$ is a multivariate Gaussian density function, $\phi(\zeta)$ will be zero as $\zeta$ tends to $\infty$. As a result, the first term Eq. (7) will be zero. Next, we analysis the derivative function of the $2nd$ term to compute $Q'(s)$ with respect to $a$,

$$\frac{\delta Q(s, h(s) + \Sigma^{1/2}\zeta)}{\delta \zeta_i} = \Sigma^{1/2} \frac{\delta Q(s,a)}{\delta a}|_{a=h(s)+\zeta\Sigma^{1/2}} \text{ where, } \delta a = \Sigma^{1/2}\delta\zeta$$

As a result, Eq. (7) can be simplified as,

$$D(s) = 0 + \int \frac{1}{\Sigma^{1/2}} \Sigma^{1/2} \frac{\delta Q(s,a)}{\delta a}|_{a=h(s)+\zeta\Sigma^{1/2}} \phi(\zeta)d\zeta$$
$$= \int Q'(s)\phi(\zeta)d\zeta \quad (8)$$

where, $Q'(s) := \frac{\delta Q(s,a)}{\delta a}|_{a=h(s)+\zeta\Sigma^{1/2}}$. Now, before advancing to proof of $D(s)$ being a Lipschitz with constant $L_D$, let us proof that $Q'(s)$ is Lipschitz continuous and bounded wtih same constant $L_D$. This step mathematically simplifies the proof for $D(s)$.

Action-value function, $Q$ is defined as, $Q(s,a) = r(s,a) + \sum_{t=1}^{\infty} \gamma^t \int r(s_t, a_t)p(s_t|a,s)\pi_h(a_t|s_t)ds_t da_t$ and its derivative with respect to $a$ is computed as,

$$Q'(s,a) = \frac{\delta r(s,a)}{\delta a} + \sum_{t=1}^{\infty} \gamma^t \int r(s_t, a_t) \frac{\delta p(s_t|a,s)}{\delta a} \pi_h(a_t|s_t)ds_t da_t \quad (9)$$

In Eq. (9), the term $\delta r(s,a)/\delta a$ is a bounded derivative by $B_{ra}$ according to Assumption 1, it will be enough to show that second derivative term in Eq. (9) $\delta p(s_t|a,s)/\delta a$ is a Lipschitz function in order to prove that $Q'(s,a)$ is a Lipschitz function as well. The transition probability distribution $p(s_t|s,a)$ can be written as,

$$p(s_t|s,a) = \int p(s_1|s,a) \prod_{u=1}^{t-1} \pi_h(a_t|s_t) p(s_{u+1}|s_u, a_u) d\mathbf{s}_{t-1} d\mathbf{a}_{t-1} \quad (10)$$

where, $d\mathbf{s}_{t-1} = (ds_1, ..., ds_{t-1})$ and $d\mathbf{a}_{t-1} = (da_1, ..., da_{t-1})$. Let us define the norm of difference of two separate $Q$-functions derivatives as follows,

$$||Q'(s,a) - Q'(s',a')|| = ||\frac{\delta r(s,a)}{\delta a} + \sum_{t=1}^{\infty} \gamma^t \int r(s_t, a_t) \frac{\delta p(s_t|a,s)}{\delta a} \pi_h(a_t|s_t)ds_t da_t - \frac{\delta r(s',a')}{\delta a}|_{a=a'}$$
$$- \sum_{t=1}^{\infty} \gamma^t \int r(s_t, a_t) \frac{\delta p(s_t|a',s')}{\delta a} \pi_h(a_t|s_t)ds_t da_t||$$
$$= ||(\frac{\delta r(s,a)}{\delta a} - \frac{\delta r(s',a')}{\delta a}) + (\sum_{t=1}^{\infty} \gamma^t \int r(s_t, a_t) \frac{\delta p(s_t|a,s)}{\delta a}|_{a=a'} \pi_h(a_t|s_t)ds_t da_t$$
$$- \sum_{t=1}^{\infty} \gamma^t \int r(s_t, a_t) \frac{\delta p(s_t|a',s')}{\delta a} \pi_h(a_t|s_t)ds_t da_t)||$$
$$= ||(\frac{\delta r(s,a)}{\delta a} - \frac{\delta r(s',a')}{\delta a}|_{a=a'}) + (\sum_{t=1}^{\infty} \gamma^t \int r(s_t, a_t)(\frac{\delta p(s_t|a,s)}{\delta a} - \frac{\delta p(s_t|a',s')}{\delta a}|_{a=a'})$$
$$\pi_h(a_t|s_t)ds_t da_t)||$$
$$\leq ||\frac{\delta r(s,a)}{\delta a} - \frac{\delta r(s',a')}{\delta a}|_{a=a'}|| + ||\sum_{t=1}^{\infty} \gamma^t \int r(s_t, a_t)(\frac{\delta p(s_t|a,s)}{\delta a} - \frac{\delta p(s_t|a',s')}{\delta a}|_{a=a'})$$
$$\pi_h(a_t|s_t)ds_t da_t||$$

We will now mention some bounds and expression which will help us to further bound and simplify the form expressed in above derived inequality. From assumption 1, we know that $|r(s,a)| < B_r$ and with bounded derivatives of reward function $||\frac{\delta r(s,a)}{\delta a} - \frac{\delta r(s',a')}{\delta a}|_{a=a'}|| \leq B_{rs}||s-s'|| + B_{ra}||a-a'||$ and using Eq. (10), we can compute the derivative of transition probability as,

$$\frac{\delta p(s_t|s,a)}{\delta a} = \int \frac{\delta p(s_1|s,a)}{\delta a} \prod_{u=1}^{t-1} \pi_h(a_t|s_t) p(s_{u+1}|s_u, a_u) d\mathbf{s}_{t-1} d\mathbf{a}_{t-1} \quad (11)$$

Therefore,

$$\frac{\delta p(s_t|a,s)}{\delta a} - \frac{\delta p(s_t|a',s')}{\delta a}\Big|_{a=a'} = \int \Big(\frac{\delta p(s_1|s,a)}{\delta a} - \frac{\delta p(s_1|s',a')}{\delta a}\Big|_{a=a'}\Big)\prod_{u=1}^{t-1}\pi_h(a_t|s_t)p(s_{u+1}|s_u,a_u)d\mathbf{s}_{t-1}d\mathbf{a}_{t-1} \quad (12)$$

Since, our assumption 2 states that $\frac{\delta p(s_1|s,a)}{\delta a} - \frac{\delta p(s_1|s',a')}{\delta a}\big|_{a=a'} \leq L_{ps}||s-s'|| + L_{pa}||a-a'||$, and both density functions $\pi_h(a_t|s_t)$ and $p(s_{u+1}|s_u,a_u)$ ultimately integrate to one, so Eq. (12) can be bounded as,

$$\frac{\delta p(s_t|a,s)}{\delta a} - \frac{\delta p(s_t|a',s')}{\delta a}\Big|_{a=a'} \leq L_{ps}||s-s'|| + L_{pa}||a-a'|| \quad (13)$$

Finally using all these above mentioned bounds and using Assumption 1, the inequality of $||Q'(s,a) - Q'(s',a')||$ can be further bounded as,

$$||Q'(s,a) - Q'(s',a')|| \leq B_{rs}||s-s'|| + B_{ra}||a-a'|| + \sum_{t=1}^{\infty}\gamma^t B_r[L_{ps}||s-s'|| + L_{pa}||a-a'||]\int \pi_h(a_t|s_t)ds_t da_t$$

$$= B_{rs}||s-s'|| + B_{ra}||a-a'|| + \frac{\gamma}{1-\gamma}B_r[L_{ps}||s-s'|| + L_{pa}||a-a'||]\int ds_t$$

$$[\text{since, }\int \pi_h(a_t|s_t)da_t = 1]$$

$$= B_{rs}||s-s'|| + B_{ra}||a-a'|| + \frac{\gamma}{1-\gamma}B_r[L_{ps}||s-s'|| + L_{pa}||a-a'||]|\mathcal{S}|$$

$$= (B_{rs} + \frac{\gamma}{1-\gamma}B_r L_{ps}|\mathcal{S}|)||s-s'|| + (B_{ra} + \frac{\gamma}{1-\gamma}B_r L_{pa}|\mathcal{S}|)||a-a'||$$

Let $L_{Qs} = B_{rs} + \frac{\gamma}{1-\gamma}B_r L_{ps}|\mathcal{S}|$ and $L_{Qa} = B_{ra} + \frac{\gamma}{1-\gamma}B_r L_{pa}|\mathcal{S}|$, so equation (15) will be further simplified as,

$$||Q'(s,a) - Q'(s',a')|| \leq L_{Qs}||s-s'|| + L_{Qa}||a-a'|| \quad (14)$$

We will continue the proof further by stating that action mapping function $h: s \to a$ is itself a Lipschitz function. Since, the function belongs to an RKHS, by exploiting the reproducing property of the kernel, the difference between $h(s)$ and $h(s')$ can be written as $h(s) - h(s') = \langle h, K(s,\cdot) - K(s',\cdot)\rangle$. Taking norm on both side yields $||h(s) - h(s')|| = ||\langle h, K(s,\cdot) - K(s',\cdot)\rangle||$. Cauchy-Schwartz inequality states that for all vectors $u$ and $v$ of an inner product space it is true that $|\langle u,v\rangle| \leq ||u||||v||$. Using the mentioned Cauchy-Schwartz inequality,

$$||h(s) - h(s')|| \leq ||h||||K(s,\cdot) - K(s',\cdot)||$$
$$or, ||h(s) - h(s')|| \leq ||h||\sqrt{K(s,s) + K(s',s') - 2K(s,s')}$$
$$or, ||h(s) - h(s')|| \leq ||h||\sqrt{2 - 2K(s,s')} \quad (15)$$

since, properties of kernel state that $K(s,s) = K(s',s') = 1$. From assumption 2, definition of kernel is $K(s,s')_{ii} = e^{-(s-s')^T\Sigma_H^{-1}(s-s')/2}$. Let us change the variable as, $u = \Sigma_H^{-1/2}(s-s')$. then $(s-s')^T\Sigma_H^{-1}(s-s') = ||u||^2$. So, the kernel function can be expressed as $K(s,s') = e^{-||u||^2/2}$. So, inequality of Eq. (15) can be written as,

$$||h(s) - h(s')|| \leq ||h||\sqrt{2}\sqrt{1 - e^{-||u||^2/2}} \quad (16)$$

Let consider, $f(u) = \sqrt{2}\sqrt{1 - e^{-||u||^2/2}}$, so expression (16) turns into $||h(s) - h(s')|| \leq ||h||f(u)$. To prove that $h(s)$ is Lipschitz function, it will be enough to prove that $f(u)$ is a Lipschitz function. The gradient of $f(u)$ yields,

$$\nabla f(u) = \frac{1}{\sqrt{2}}\frac{e^{-||u||^2/2}u}{\sqrt{1 - e^{-||u||^2/2}}} \quad (17)$$

It seems from the Eq. (17) that the function $f(u)$ might not be bounded at $u = 0$, as limit of denominator goes to zero. But this is not entirely true which can be shown by a Taylor's second order approximation of denominator term,

$$1 - e^{-||u||^2/2} = \frac{1}{2}||u||^2 + o(||u||^2)$$

where for any function $o(||u||^2)$, $\lim_{||u||\to 0} o(||u||^2)/||u||^2 = 0$. Thus,

$$\lim_{||u||\to 0}||\nabla f(u)|| = \lim_{||u||\to 0}\frac{1}{\sqrt{2}}\frac{e^{-||u||^2/2}u}{\sqrt{1/2||u||^2}}$$

$$= lim_{||u||\to 0}\frac{e^{-||u||^2/2}u}{u}$$

$$= 1$$

Besides, the gradient of $||\nabla f(u)||$ is always differentiable except at $u = 0$ and it never attains the value 0 except when $||u|| \to \infty$. Thus, all critical points of $||\nabla f(u)||$) lies in infinity. On contrary, from (17) $\lim_{||u||\to\infty}||\nabla f(u)|| = 0$, so the critical points at infinity is a minimum. Thus, the maximum norm of $\nabla f(u)$ is attained when $u = 0$ and it takes the value 1. Thus, $f(u)$ is Lipschitz with constant 1. Using $f(0) = 0$ to bound, we get $|f(u)| \leq ||u|| = ||\Sigma_H^{-1/2}(s-s')|| \leq \lambda_{min}\Sigma_H^{-1/2}||s-s'||$. Using

this Lipschitz continuity and bounds, we can clearly state from (16) that $h(s)$ is Lipschitz with constant $L_h := ||h||\lambda_{min}\Sigma_H^{-1/2}$. As a result,

$$\begin{aligned}
||h(s) - h(s')|| &\leq ||h||f(u) \\
\text{or, } ||h(s) - h(s')|| &\leq ||h||\lambda_{min}\Sigma_H^{-1/2}||s - s'|| \\
\text{or, } |h(s) - h(s')| &\leq L_h||s - s'||
\end{aligned} \quad (18)$$

Finally, let's rewrite the Eq. (14) with the action mapping function such as,

$$\begin{aligned}
||Q'(s,a) - Q'(s',a')|| &\leq L_{Qs}||s - s'|| + L_{Qa}||h(s) - h(s')|| \\
\text{or, } ||Q'(s,a) - Q'(s',a')|| &\leq L_{Qs}||s - s'|| + L_{Qa}L_h||s - s'|| \\
\text{or, } ||Q'(s,a) - Q'(s',a')|| &\leq (L_{Qs} + L_{Qa}L_h)||s - s'|| \\
\text{or, } ||Q'(s,a) - Q'(s',a')|| &\leq L_D||s - s'||
\end{aligned} \quad (19)$$

where, $L_D := L_{Qs} + L_{Qa}L_h$. Since, $Q(s)$ is Lipschitz with constant $L_D$, then $D(s)$ is also Lipschitz with constant $L_D$ following Eq. (8). To proof that $D(s)$ is also bounded, we need to consider it's norm and use expression (6) to set the possible upper bound as,

$$||D(s)|| \leq \mathbb{E}||Q(s, h(s) + \Sigma^{1/2}\zeta)\zeta|| \quad (20)$$

From equation (44), $|Q(s,a) < B_r/(1-\gamma)|$. So the above inequality can be further bounded as,

$$||D(s)|| \leq \frac{B_r}{1-\gamma}\mathbb{E}||\zeta|| = B_D \quad (21)$$

where, $B_D := \frac{B_r}{1-\gamma}\mathbb{E}||\zeta|| = \frac{B_r}{1-\gamma}\mathbb{E}||\Sigma^{-1/2}(a - h(s))||$. To finalize the proof of our Lemma 1, we can write the difference between two instance of $D(s)\rho_{s_0}(s)$ using triangular inequality as,

$$\begin{aligned}
||D(s)\rho_{s_0}(s) - D(s')\rho_{s_0}(s')|| &\leq ||D(s)\rho_{s_0}(s) - D(s)\rho_{s_0}(s')|| + ||D(s)\rho_{s_0}(s') - D(s')\rho_{s_0}(s')|| \\
\text{or, } ||D(s)\rho_{s_0}(s) - D(s')\rho_{s_0}(s')|| &\leq D(s)||\rho_{s_0}(s) - \rho_{s_0}(s')|| + \rho_{s_0}(s')||D(s) - D(s')||
\end{aligned} \quad (22)$$

Using from (21), $D(s) \leq B_D$, from (19) $||Q'(s,a) - Q'(s',a')|| \leq L_D||s - s'||$ or $||D(s) - D(s')|| \leq L_D||s - s'||$, from (5) $|\rho_0(s) - \rho_0(s')| \leq L_p||s - s'||$ and from assumption 2, $\rho_{s_0}(s) \leq B_1$, the inequality (22) can be further bounded by,

$$\begin{aligned}
||D(s)\rho_{s_0}(s) - D(s')\rho_{s_0}(s')|| &\leq B_D L_p||s - s'|| + B_1 L_D||s - s'|| \\
\text{or, } ||D(s)\rho_{s_0}(s) - D(s')\rho_{s_0}(s')|| &\leq L||s - s'||
\end{aligned} \quad (23)$$

where, $L := B_D L_p + B_1 L_D$. So we can finally state $D(s)\rho_{s_0}(s)$ is a bounded function as $D(s)\rho_{s_0}(s) \leq B_D B_1 = B$ and follows Lipschitz continuity with constant L.

**Theorem 1:**

For every $\varepsilon > 0$ and every $h \in \mathcal{H}$ satisfying below mentioned condition for maintaining the maximum width of kernel $K(s,s')$,

$$\sqrt{nm}||\Sigma_H||LB|\mathcal{S}| \leq \varepsilon/2$$

if the norm of the gradient of the utility function maintains the bounded condition of $||\nabla_h J_{s_0}(h,.)||_\mathcal{H}^2 \geq \varepsilon$ which means that current functional $h$ is $\varepsilon$ neighborhood out of the optimal solution for current observed state $s_0$ then we can state that,

$$\langle \nabla_h J_{s_0}(h,.)|_{s=s_0}, \nabla_h J_{s_0}(h,.)|_{s=s'}\rangle_\mathcal{H} \geq Z\frac{\varepsilon}{2}$$

where $Z := \sqrt{2\pi \, det(\Sigma_H)}$ and $\Sigma_H$ is the covariance matrix of Gaussian Kernel $K_{\Sigma_H}(s,s')$. The above inner-product based similarity evaluation checks whether gradients for future states $s'$ is following an ascent direction with respect the gradient calculated at initial state $s_0$ to maximize long term utility and guarantee optimal action execution.

*Proof:* The inner product $\langle \nabla_h J_{s_0}(h,.)|_{s=s_0}, \nabla_h J_{s_0}(h,.)|_{s=s'}\rangle_\mathcal{H}$ can be represented by their respective integral based policy gradient representation in the following way,

$$I_{\Sigma_H} = \int D(s)^T \rho_{s_0}(s) K_{\Sigma_H}(s,s') D(s') \rho_{s_0}(s') ds ds' \quad (24)$$

The change is state space representation can be written as, $u = s' - s$. Now, replacing the $s'$ by $s + u$ in Eq. (24) gives,

$$I_{\Sigma_H} = \int D(s)^T \rho_{s_0}(s) K_{\Sigma_H}(s, s+u) D(s+u) \rho_{s_0}(s+u) ds du \quad (25)$$

$$= Z \int D(s)^T \rho_{s_0}(s) \frac{K_{\Sigma_H}(s, s+u)}{Z} D(s+u) \rho_{s_0}(s+u) ds du \quad (26)$$

$$= Z \int D(s)^T \rho_{s_0}(s) G(u; 0, \Sigma_H) D(s+u) \rho_{s_0}(s+u) ds du \quad (27)$$

Here, $Z := \sqrt{2\pi \, det(\Sigma_H)}$ is the normalizing factor and $G(u; 0, \Sigma_H)$ is a Gaussian probability density function with 0 mean and $\Sigma_H$ covariance, $u$ is sampled from $G$. The integral expression with respect to $u$ in Eq. (27) can be written as the expectation of $D(s+u)\rho_{s_0}(s+u)$ with respect to $u$ sampled from a Normal distribution in following manner,

$$I_{\Sigma_H} = Z \int D(s)^T \rho_{s_0}(s) \mathbb{E}_{u \sim \mathcal{N}(0, \Sigma_H)}[D(s+u)\rho_{s_0}(s+u)] ds \quad (28)$$

Now, according to [[3], Theorem 1], if any function $f(x)$ is Lipschitiz continuous function with constant L and $f_\mu(x)$ is a Gaussian approximation with smoothness parameter $\mu$, then,

$$|f_\mu(x) - f(x)| \leq \sqrt{n}\mu L$$

Here, $f_\mu(x) = \frac{1}{K}\int f(x+\mu u)e^{-\frac{1}{2}||u||^2}du$ and $n = \frac{1}{K}\int ||u||^2 e^{-\frac{1}{2}||u||^2}du$. This Gaussian approximation is calculated in a scalar-valued space.

In our problem statement, kernel $K_{\Sigma_H}(s,s')$ is matrix valued Gaussian Kernel with covariance matrix $\Sigma_H$ such that for $i = 0, 1..., m$, we have that

$$K_{\Sigma_H}(s,s')_{ii} = e^{-(s-s')^T \Sigma_H^{-1}(s-s')/2}$$

When $i \neq j$, then $K_{\Sigma_H}(s,s')_{ij} = 0$ for all $j = 0,..,m$. From Eq. (28), the expectation based term can be expressed as,

$$\mathbb{E}_{u \sim \mathcal{N}(0,\Sigma_H)}[D(s+u)\rho_{s_0}(s+u)] = \frac{1}{Z}\int D(s+u)\rho_{s_0}(s+u)K_{\Sigma_H}(s,s+u)du \tag{29}$$

$$= \frac{1}{Z}\int D(s+u)\rho_{s_0}(s+u)e^{-(-u)^T \Sigma_H^{-1}(-u)}du \tag{30}$$

According to Lemma 1, we have $D(s)\rho_{s_0}(s)$ is Lipschitz with constant L and Eq. (30) implies that expectation of $D(s+u)\rho_{s_0}(s+u)$ is a Gaussian approximation with covariance parameter $\Sigma_H$, then by virtue of [[3], Theorem 1], we can state that,

$$|\mathbb{E}_{u \sim \mathcal{N}(0,\Sigma_H)}[D(s+u)\rho_{s_0}(s+u)] - D(s)\rho_{s_0}(s)| \leq \sqrt{nm}||\Sigma_H||L \tag{31}$$

From the properties of absolute value, we know the inequality based relationship that says $|A - B| \geq |A| - |B|$. Using this property, the inequality of Eq. (31) can be further bounded by,

$$|\mathbb{E}_{u \sim \mathcal{N}(0,\Sigma_H)}[D(s+u)\rho_{s_0}(s+u)]| - |D(s)\rho_{s_0}(s)| \leq \sqrt{nm}||\Sigma_H||L \tag{32}$$

$$or, \ |\mathbb{E}_{u \sim \mathcal{N}(0,\Sigma_H)}[D(s+u)\rho_{s_0}(s+u)]| \leq \sqrt{nm}||\Sigma_H||L + |D(s)\rho_{s_0}(s)| \tag{33}$$

$$or, \ \mathbb{E}_{u \sim \mathcal{N}(0,\Sigma_H)}[D(s+u)\rho_{s_0}(s+u)] \leq \sqrt{nm}||\Sigma_H||L + |D(s)\rho_{s_0}(s)| \tag{34}$$

Using the inequality of (34), the expression of $I_{\Sigma_H}$ in Eq. (28) can be lower bounded by,

$$I_{\Sigma_H} \geq Z \int D(s)^T \rho_{s_0}(s)\big[|D(s)\rho_{s_0}(s)| - \sqrt{nm}||\Sigma_H||L\big]ds \tag{35}$$

$$\geq Z \int ||D(s)||^2 \rho_{s_0}^2(s)ds - Z\sqrt{nm}||\Sigma_H||L\int |D(s)\rho_{s_0}(s)|ds \tag{36}$$

We can define $|\mathcal{S}|$ as the measure of the set $\mathcal{S}$. Under Lemma 1, $D(s)\rho_{s_0}(s)$ is bounded by a constant B, so the inequality of Eq. (36) can be further lower bounded by,

$$I_{\Sigma_H} \geq Z \int ||D(s)||^2 \rho_{s_0}^2(s)ds - \sqrt{nm}||\Sigma_H||ZLB|\mathcal{S}| \tag{37}$$

By using the integral based policy gradient expression, $||\nabla_h J_{s_0}(h,.)||_{\mathcal{H}}^2 = \int ||D(s)||^2 \rho_{s_0}^2(s)ds$. From the statement of theorem 1, $||\nabla_h J_{s_0}(h,.)||_{\mathcal{H}}^2 \geq \varepsilon$. So, $\int ||D(s)||^2 \rho_{s_0}^2(s)ds \geq \varepsilon$. So the inequality in expression (37), can be further bounded by,

$$I_{\Sigma_H} \geq Z\varepsilon - \sqrt{nm}||\Sigma_H||ZLB|\mathcal{S}| \tag{38}$$

By using the condition of maximum kernel width,

$$I_{\Sigma_H} \geq Z\varepsilon - Z\frac{\varepsilon}{2} \tag{39}$$

$$\geq Z\frac{\varepsilon}{2} \tag{40}$$

This concludes the proof of theorem 1.

**Lemma 2:** Let $\Gamma(\cdot)$ is a Gamma function and define,

$$\sigma = \frac{B_r}{(1-\gamma)^2}\frac{1}{\Sigma^{1/2}}\Big(4\frac{\Gamma(2+\frac{p}{2})}{\Gamma(\frac{p}{2})}\Big)^{1/4}$$

Then the following bound holds true,

$$\mathbb{E}[||\nabla_h J_{s_0}(h,.)||^2] \leq \sigma^2 \quad \text{and} \quad \mathbb{E}[||\nabla_h J_{s_0}(h,.)||^3] \leq \sigma^3$$

*Proof:* This proof is followed by [[4], Lemma 6] to some extent.
The expression of calculating the gradient is given by,

$$\nabla_h \hat{J}_{s_0}(h, \cdot) = \frac{1}{1-\gamma}\hat{Q}(s_t, a_t; h)K(s_t, \cdot)\Sigma^{-1}(a_t - h(s_t)) \tag{41}$$

The action-value function estimate is defined as,

$$\hat{Q}(s, a; h) = \mathbb{E}\Big[\sum_{t=0}^{\infty} \gamma^t r(s_t, a_t)|h, s_0 = s, a_0 = a\Big] \tag{42}$$

Taking norm on both sides of Eq.(42) yields,

$$|\hat{Q}(s,a;h)| = |\mathbb{E}\big[\sum_{t=0}^{\infty}\gamma^t r(s_t,a_t)|h, s_0=s, a_0=a\big]|$$

$$\leq \mathbb{E}|\big[\sum_{t=0}^{\infty}\gamma^t r(s_t,a_t)|h, s_0=s, a_0=a\big]| \quad (43)$$

According to assumption 1, $|r(s_t,a_t)| \leq B_r \ \forall (s,a) \in \mathcal{S}\times\mathcal{A}$. Using this bound, inequality of (43) can be further bounded by,

$$|\hat{Q}(s,a;h)| \leq B_r \sum_{t=0}^{\infty} \gamma^t$$
$$= \frac{B_r}{1-\gamma} \quad (44)$$

Now, let's start the proof of the lemma by taking cube of the norm of gradient expression,

$$||\nabla_h \hat{J}_{s_0}(h,\cdot)||^3 = \frac{1}{(1-\gamma)^3}||\hat{Q}(s_t,a_t;h)K(s,\cdot)\Sigma^{-1}(a_t - h(s_t))||^3$$
$$\leq \frac{1}{(1-\gamma)^3}||\hat{Q}(s_t,a_t;h)||^3 ||K(s,\cdot)||^3 ||\Sigma^{-1}(a_t - h(s_t))||^3 \quad (45)$$

By definition of RKHS, $||K(s,\cdot)|| = 1$. Using the bounds of (44), the inequality of (45) can be further bounded by,

$$||\nabla_h \hat{J}_{s_0}(h,\cdot)||^3 \leq \frac{B_r^3}{(1-\gamma)^6}||\Sigma^{-1}(a_t - h(s_t))||^3 \quad (46)$$

By taking expectation on both side of (46) and applying monotonicity of expectation,

$$\mathbb{E}\big[||\nabla_h \hat{J}_{s_0}(h,\cdot)||^3\big] \leq \frac{B_r^3}{(1-\gamma)^6}\mathbb{E}\big[||\Sigma^{-1}(a_t - h(s_t))||^3\big] \quad (47)$$

The further steps will to be simplify the expectation based term of (47) by using two properties of Chi-Squared distribution.

Property 1: If $Z_1, Z_2, ..., Z_k$ are independent standard normal random variables, then $\sum_{i=1}^{k}(Z_i - \bar{Z})^2 \sim \mathcal{X}_p^2$ where, $\mathcal{X}_p^2$ is a Chi-Squared with parameter p and $\bar{Z}$ is the mean.

Now, $||\Sigma^{-1/2}(a_t - h(s_t))||^2$ can be interpreted as Chi-Squared with parameter $p$, since $h(s_t)$ is the mean of the Gaussian policy and the term is a sum of squares of all normal random variables. Then, according to property 1 mentioned earlier and by applying Jensen's inequality,

$$\mathbb{E}\big[||\Sigma^{-1}(a_t - h(s_t))||^3\big] = \mathbb{E}\big[||\Sigma^{-1/2}\Sigma^{-1/2}(a_t - h(s_t))||^3\big]$$
$$\leq \frac{1}{(\Sigma^{1/2})^3}\mathbb{E}\big[\big[||\Sigma^{-1/2}(a_t - h(s_t))||^2\big]^{3/2}\big]$$
$$= \frac{1}{(\Sigma^{1/2})^3}\mathbb{E}\big[\mathcal{X}_p^{3/2}\big]$$
$$= \frac{1}{(\Sigma^{1/2})^3}\mathbb{E}\big[\mathcal{X}_p^2\big]^{3/4} \quad (48)$$

Property 2: Non-central moments of Chi-Squared distribution follows the following property,

$$\mathbb{E}\big[\mathcal{X}_p^m\big] = 2^m \frac{\Gamma(2+\frac{p}{2})}{\Gamma(\frac{p}{2})}$$

Using property 2,

$$\mathbb{E}\big[\mathcal{X}_p^2\big] = 4\frac{\Gamma(2+\frac{p}{2})}{\Gamma(\frac{p}{2})} \quad (49)$$

Using the value from Eq. (49), inequality (48) can be expressed as,

$$\mathbb{E}\big[||\Sigma^{-1}(a_t - h(s_t))||^3\big] \leq \frac{1}{(\Sigma^{1/2})^3}\big(4\frac{\Gamma(2+\frac{p}{2})}{\Gamma(\frac{p}{2})}\big)^{3/4} \quad (50)$$

Let us define, $\sigma = \frac{B_r}{(1-\gamma)^2}\frac{1}{\Sigma^{1/2}}\big(4\frac{\Gamma(2+\frac{p}{2})}{\Gamma(\frac{p}{2})}\big)^{1/4}$. Then using the bound deduced in inequality (50), expression of (47) can be further bounded by,

$$\mathbb{E}\big[||\nabla_h \hat{J}_{s_0}(h,\cdot)||^3\big] \leq \frac{B_r^3}{(1-\gamma)^6}\frac{1}{(\Sigma^{1/2})^3}\big(4\frac{\Gamma(2+\frac{p}{2})}{\Gamma(\frac{p}{2})}\big)^{3/4}$$
$$= \sigma^3 \quad (51)$$

The analogous proof can be made for $\mathbb{E}\big[||\nabla_h \hat{J}_{s_0}(h,\cdot)||^2\big] \leq \sigma^2$.

**Lemma 3:**

To construct the sub-martingale for theorem 2, a lower bound on the expectation of random variables $J_{s_0}(h_{k+1})$ conditioned on the sigma filed $\mathcal{F}_k$ is provided in this lemma.

$$\mathbb{E}\big[J_{s_0}(h_{k+1})|\mathcal{F}_k\big] \geq J_{s_0}(h_k) - \alpha_k^2 C_1 - \alpha_k^3 C_2 + \alpha_k \langle \nabla_h J_{s_0}(h,.)|_{s=s_0}, \nabla_h J_{s_0}(h,.)|_{s=s'} \rangle_{\mathcal{H}}$$

where, $\alpha_k$ is the step-size of the algorithm. $C_1 := L_1 \sigma^2$ and $C_1 := L_2 \sigma^3$.

$$L_1 = B_r \frac{(1-\gamma) + p(1+\gamma)}{\lambda_{min} \Sigma (1-\gamma)^2}$$
$$L_2 = B_r \frac{(1+p)\sqrt{p}}{\lambda_{min} \Sigma^{3/2} (1-\gamma)^3}$$
$$\sigma = \frac{B_r}{(1-\gamma)^2} \frac{1}{\Sigma^{1/2}} \left(4 \frac{\Gamma(2+\frac{p}{2})}{\Gamma(\frac{p}{2})}\right)^{1/4}$$

*Proof:* Let's start by writing the Taylor expansion of $J_{s_0}(h_{k+1})$ around $h_k$,

$$J_{s_0}(h_{k+1}) = J_{s_0}(h_k) + \langle \nabla_h J_{s_0}(h_{k+1}), h_{k+1} - h_k \rangle_{\mathcal{H}}$$

By adding and subtracting $\langle \nabla_h J_{s_0}(h_k), h_{k+1} - h_k \rangle_{\mathcal{H}}$ in the previous equation yields,

$$J_{s_0}(h_{k+1}) = J_{s_0}(h_k) + \langle \nabla_h J_{s_0}(h_{k+1}), h_{k+1} - h_k \rangle_{\mathcal{H}} + \langle \nabla_h J_{s_0}(h_k), h_{k+1} - h_k \rangle_{\mathcal{H}} - \langle \nabla_h J_{s_0}(h_k), h_{k+1} - h_k \rangle_{\mathcal{H}}$$
$$= J_{s_0}(h_k) + \langle \nabla_h J_{s_0}(h_{k+1}) - \nabla_h J_{s_0}(h_k), h_{k+1} - h_k \rangle_{\mathcal{H}} + \langle \nabla_h J_{s_0}(h_k), h_{k+1} - h_k \rangle_{\mathcal{H}} \quad (52)$$

According to [[4], Lemma 5], the gradient of the expected discounted reward satisfies,

$$||\nabla_h J_{s_0}(h_1) - \nabla_h J_{s_0}(h_2)||_{\mathcal{H}} \leq L_1 ||h_1 - h_2||_{\mathcal{H}} + L_2 ||h_1 - h_2||_{\mathcal{H}}^2$$

This result can be extended as,

$$||\nabla_h J_{s_0}(h_{k+1}) - \nabla_h J_{s_0}(h_k)||_{\mathcal{H}} \leq L_1 ||h_{k+1} - h_k||_{\mathcal{H}} + L_2 ||h_{k+1} - h_k||_{\mathcal{H}}^2 \quad (53)$$

Using this bounded form of (53), Eq. (52) can be lower bounded by,

$$J_{s_0}(h_{k+1}) \geq J_{s_0}(h_k) - \langle (L_1 ||h_{k+1} - h_k||_{\mathcal{H}} + L_2 ||h_{k+1} - h_k||_{\mathcal{H}}^2), h_{k+1} - h_k \rangle_{\mathcal{H}} + \langle \nabla_h J_{s_0}(h_k), h_{k+1} - h_k \rangle_{\mathcal{H}} \quad (54)$$

By applying Cauchy-Schwartz inequality in first inner-product term, the inequality of (54) can be further bounded by,

$$J_{s_0}(h_{k+1}) \geq J_{s_0}(h_k) - L_1 ||h_{k+1} - h_k||_{\mathcal{H}}^2 - L_2 ||h_{k+1} - h_k||_{\mathcal{H}}^3 + \langle \nabla_h J_{s_0}(h_k), h_{k+1} - h_k \rangle_{\mathcal{H}} \quad (55)$$

Now, taking the expectation with respect to sigma field $\mathcal{F}_k$ and combining the monotonicity and linearity of expectation the inequality (55) can be expressed as,

$$\mathbb{E}[J_{s_0}(h_{k+1})|\mathcal{F}_k] \geq J_{s_0}(h_k) - L_1 \mathbb{E}[||h_{k+1} - h_k||_{\mathcal{H}}^2 |\mathcal{F}_k] - L_2 \mathbb{E}[||h_{k+1} - h_k||_{\mathcal{H}}^3 |\mathcal{F}_k] + \langle \nabla_h J_{s_0}(h_k), \mathbb{E}[h_{k+1} - h_k |\mathcal{F}_k] \rangle_{\mathcal{H}} \quad (56)$$

The theory of gradient ascent algorithm states that

$$h_{k+1} = h_k + \alpha_k \nabla_h J_{s_0}(h_k, \cdot)|_{s=s'}$$
$$or, h_{k+1} - h_k = \alpha_k \nabla_h J_{s_0}(h_k, \cdot)|_{s=s'}$$

So,

$$L_1 \mathbb{E}[||h_{k+1} - h_k||_{\mathcal{H}}^2 |\mathcal{F}_k] = \alpha_k^2 L_1 [||\nabla_h J_{s_0}(h_k, \cdot)|_{s=s'}||_{\mathcal{H}}^2 |\mathcal{F}_k]$$
$$\leq \alpha_k^2 L_1 \sigma^2 \quad \text{[by applying results from Lemma 2]}$$
$$= \alpha_k^2 C_1 \quad (57)$$

In the same way, we can deduce that,

$$L_2 \mathbb{E}[||h_{k+1} - h_k||_{\mathcal{H}}^3 |\mathcal{F}_k] \leq \alpha_k^3 C_2 \quad (58)$$

The fourth term on the RHS of inequality (56) can be further simplified as,

$$\langle \nabla_h J_{s_0}(h_k), \mathbb{E}[h_{k+1} - h_k |\mathcal{F}_k] \rangle_{\mathcal{H}} = \langle \nabla_h J_{s_0}(h_k), \mathbb{E}[\alpha_k \nabla_h J_{s_0}(h_k, \cdot)|_{s=s'} |\mathcal{F}_k] \rangle_{\mathcal{H}}$$
$$= \alpha_k \langle \nabla_h J_{s_0}(h_k), \nabla_h J_{s_0}(h_k, \cdot)|_{s=s'} \rangle_{\mathcal{H}} \quad (59)$$

Finally, with the bounds introduced in (57), (58) and equation (59), the whole expression of (56) can be further bounded by,

$$\mathbb{E}[J_{s_0}(h_{k+1})|\mathcal{F}_k] \geq J_{s_0}(h_k) - \alpha_k^2 C_1 - \alpha_k^3 C_2 + \alpha_k \langle \nabla_h J_{s_0}(h_k), \nabla_h J_{s_0}(h_k, \cdot)|_{s=s'} \rangle_{\mathcal{H}}$$

This concludes the proof of Lemma 3.

Convergence Analysis: Let's define $(\Omega, \mathcal{F}, P)$ be a probability space and the following increasing sequence of sigma-algebras $\{\emptyset, \Omega\} = \mathcal{F}_0 \subset \mathcal{F}_1 ... \subset \mathcal{F}_k ... \subset \mathcal{F}_\infty \subset \mathcal{F}$. Here, each $\mathcal{F}_k$ is generated by random variables $h_0, ..., h_k$. Also, let's define dynamic regret as,

$$R_k = J_{s_0}(h_k^*) - J_{s_0}(h_k)$$

**Theorem 2:**
For algorithmic step-size $\alpha_k > 0$ such that

$$\alpha_k \leq \frac{\sqrt{C_1^2 + 4C_2 Z \frac{\varepsilon}{2}} - C_1}{2C_2}$$

with Gaussian kernel maintaining the condition of maximum kernel width and results from theorem 1, the sequence of action mapping functions turn $R_k$ into a sub-martingale and according to Martingale Convergence Theorem, $\mathbb{E}[R^*] = \lim_{k \to \infty} R_k$

*Proof:* Dynamic Regret is defined as,

$$R_k = J_{s_0}(h_k^*) - J_{s_0}(h_k)$$

We need to show that $R_k$ is a non-negative sub-martingale. Since, $J_{s_0}(h_k^*)$ maximizes $J_{s_0}(h_k)$, then $R_k$ is always non-negative. Again, since, $J_{s_0}(h_k)$ is $\mathcal{F}_k$–measurable, then $R_k$ is also $\mathcal{F}_k$–measurable. We need to prove now that, $\mathbb{E}[R_{k+1}|\mathcal{F}_k] \leq R_k$. Using the result of Lemma 3, we can set an upper-bound on $\mathbb{E}[R_{k+1}|\mathcal{F}_k]$ as,

$$\mathbb{E}[R_{k+1}|\mathcal{F}_k] \leq R_k + \alpha_k^2 C_1 + \alpha_k^3 C_2 - \alpha_k \langle \nabla_h J_{s_0}(h_k), \nabla_h J_{s_0}(h_k, \cdot)|_{s=s'} \rangle_{\mathcal{H}} \tag{60}$$

From the result proved in Theorem 1, the previous inequality can be further bounded by,

$$\begin{aligned}
\mathbb{E}[R_{k+1}|\mathcal{F}_k] &\leq R_k + \alpha_k^2 C_1 + \alpha_k^3 C_2 - \alpha_k Z \frac{\varepsilon}{2} \\
&= R_k + \alpha_k [\alpha_k C_1 + \alpha_k^2 C_2 - Z \frac{\varepsilon}{2}] \\
&= R_k + \alpha_k \beta(\alpha_k)
\end{aligned} \tag{61}$$

where, $\beta(\alpha_k) = \alpha_k C_1 + \alpha_k^2 C_2 - Z \frac{\varepsilon}{2}$. This expression will be always negative when $\alpha_k$ maintains the bound,

$$\alpha_k \leq \frac{\sqrt{c_1^2 + 4C_2 Z \frac{\varepsilon}{2}} - C_1}{2C_2}$$

With the term $\beta(\alpha_k)$ being always negative, then $R_k$ is a non-negative sub-martingale. So, $\mathbb{E}[R_{k+1}|\mathcal{F}_k] \leq R_k$ is true. So, we conclude that with adaptively revisiting the policy udpate at perfect interruption point with the help emerging from the inner product based ascent direction check[proved in Theorem 1], our proposed approach guarantees convergence to the optimal solution.



\end{document}